\documentclass[11pt]{article}

\usepackage{acl}
\usepackage{times}
\usepackage{latexsym}
\usepackage[T1]{fontenc}
\usepackage[utf8]{inputenc}
\usepackage{microtype}
\usepackage{makecell}
\usepackage{inconsolata}
\usepackage{tabularx}
\usepackage{array}
\usepackage{graphicx}
\usepackage{amsmath}
\usepackage{amssymb}
\usepackage{booktabs}
\usepackage{multirow}

\newcolumntype{Y}{>{\centering\arraybackslash}X}
\title{DeGMatch: Target-Aware Reasoning Data Selection via Deconfounded Gradient Matching}

\author{
\textbf{Changho Lee}$^{1}$ \quad
\textbf{Janghoon Han}$^{2}$ \quad
\textbf{Rajdeep Mukherjee}$^{1}$ \quad
\textbf{Yeda Song}$^{1}$ \\
\textbf{Lening Nick Cui}$^{1}$ \quad
\textbf{Xinya Du}$^{2}$ \quad
\textbf{Honglak Lee}$^{1,3}$ \\[2pt]
$^{1}$University of Michigan \quad
$^{2}$University of Texas at Dallas \quad
$^{3}$LG AI Research \\[2pt]
\texttt{changhol@umich.edu}
}
\begin{document}

\maketitle

\begin{abstract}
Reasoning post-training on long trajectories has substantially improved the reasoning capabilities of large language models, but scaling such training to large datasets is increasingly expensive. We study \textit{target-aware reasoning data selection}, where a small reference set specifies a target capability and guides the selection of training trajectories for specialization. We find that existing target-aware methods can exhibit limited target discrimination on long reasoning trajectories, while dominant population-level structure in candidate gradients can obscure finer target-dependent variation during gradient matching. To address these limitations, we introduce \textbf{De}confounded \textbf{G}radient \textbf{Match}ing (\textbf{DeGMatch}), which first constructs a target-relevance gate and then reranks candidates using model-specific gradients. DeGMatch centers gradients using the candidate-population mean and removes leading principal directions before target matching, reducing the influence of dominant population-level structure and improving target discrimination. Across mathematics, science, and code, DeGMatch outperforms strong heuristic, target-aware, and reasoning-specific data selection baselines, while remaining effective across selection budgets, model families, and scales. Our code is publicly available at \url{https://github.com/CHLee0801/degmatch}.
\end{abstract}

\section{Introduction}
\label{sec:introduction}

\begin{figure}[t]
    \centering
    \includegraphics[width=\columnwidth]{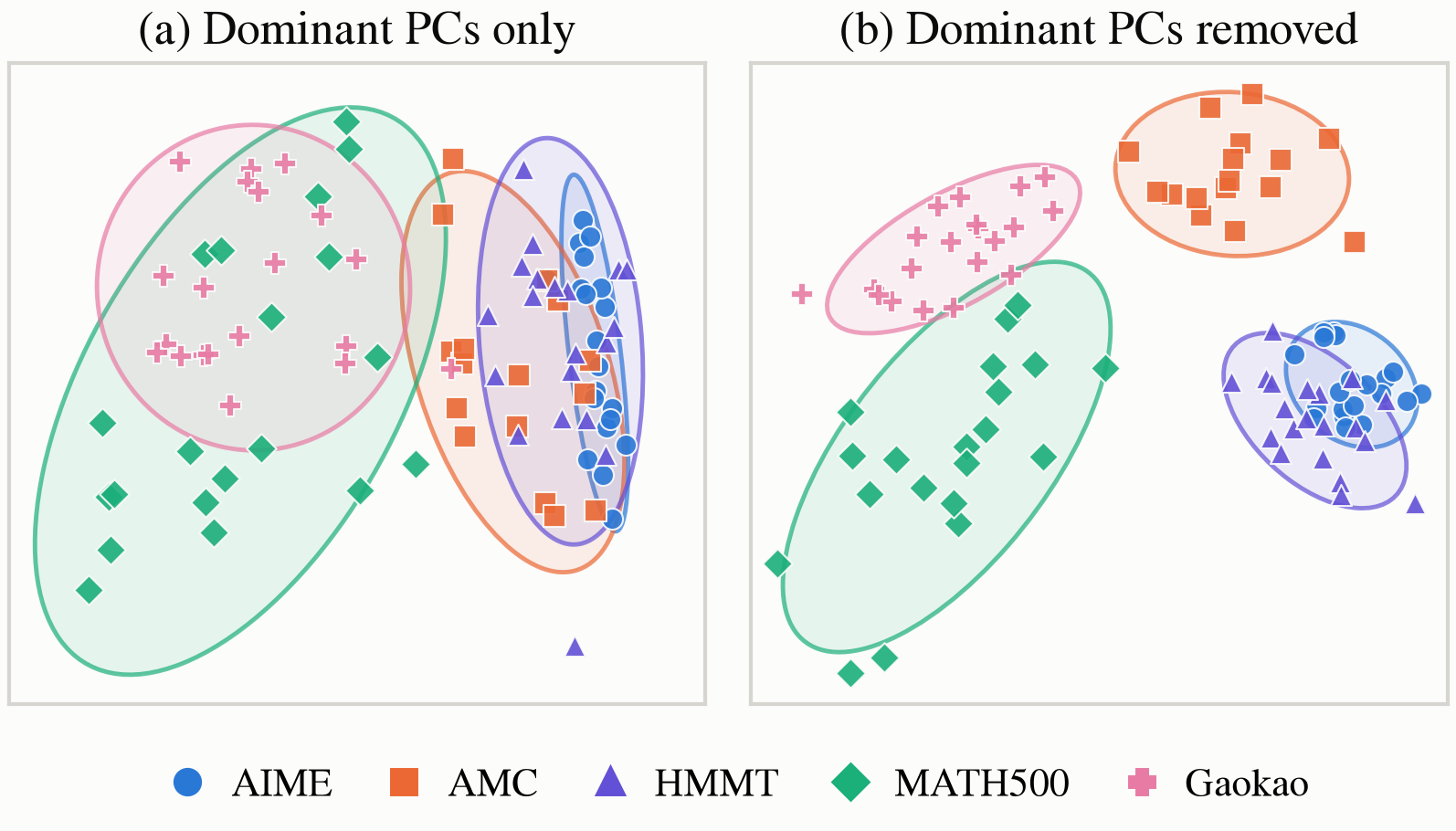}
    \caption{
    \textbf{Dominant gradient structure obscures fine-grained target variation.}
    Each point represents a reference subset positioned by the candidate ranking it induces.
    \textbf{(a)} Rankings using only the top-50 principal-component subspace.
    \textbf{(b)} Rankings after removing the population mean and top 50 principal components.
    The corresponding five-way target-identification accuracy, computed in the original ranking space, increases from 56\% to 71\%.
    See Appendix~\ref{app:reference_subset_structure} for details.
    }
    \label{fig:ranking_geometry}
    \vspace{-10pt}
\end{figure}
Recent advances in large language models have yielded substantial improvements in complex reasoning capabilities~\citep{guo2025deepseekr1,yang2025qwen3}. Reasoning post-training has emerged as a standard approach to enhancing these capabilities through training on long reasoning trajectories~\citep{guha2026openthoughts,chen2025acereason,yuan2025naturalreasoning,zhao2025distilledreasoning}. However, the growing scale of reasoning datasets and the length of individual trajectories make full-data training increasingly computationally expensive. This has motivated growing interest in data-efficient reasoning post-training, where carefully selected subsets of reasoning trajectories can achieve competitive performance with substantially less training data~\citep{yang2026select2reason,li2025naturalthoughts,yang2026informativealignment,yu2026lark,li2026grace}.

While data-efficient selection seeks compact subsets that are broadly useful for post-training, many real-world applications require specialization to a particular downstream task or capability rather than broad, general-purpose competence~\citep{xia2024less}. For such specialization, training examples that benefit general reasoning performance may not be those that best support the target capability. This motivates \emph{target-aware data selection}, which aims to identify training examples that most effectively support a specified downstream objective. 
Prior work has explored this paradigm in instruction tuning and general data-selection settings, using target similarity or distribution matching to identify target-relevant examples, and gradient-based signals to estimate their training utility for a specified target~\citep{liu2024tsds,ivison2025largescale,lee2024instructionmatters,xia2024less,feng2024tarot,min2026gist}.

However, applying gradient-based matching to long reasoning trajectories reveals an important limitation. We find that candidate gradients are dominated by a small number of population-level directions: across five mathematical target-conditioned candidate populations, the top 50 principal components explain 86.7--89.4\% of the centered-gradient variance. The leading principal subspaces also remain substantially aligned across candidate populations induced by different targets. As shown in Figure~\ref{fig:ranking_geometry}, rankings formed from the dominant subspace exhibit limited fine-grained target separation, whereas removing these dominant directions yields more distinguishable target-dependent ranking structure. These results suggest that dominant population-level gradient structure can obscure finer target-dependent variation during gradient matching.

Motivated by this analysis, we introduce \textbf{DeGMatch}, which suppresses dominant population-level structure in candidate gradients to better expose target-dependent variation during gradient matching. DeGMatch first restricts selection to a target-relevant candidate region. Within this region, it computes gradients with respect to the model being post-trained and, inspired by common-component removal in representation learning~\citep{mu2017allbuttop}, centers gradients using the candidate-population mean and removes the leading principal directions estimated from the candidate gradients before target matching. DeGMatch then scores candidates using directional alignment with target gradients together with their gradient magnitude in the target-gradient subspace.

Empirically, DeGMatch consistently outperforms strong heuristic, target-aware, and reasoning-specific data selection baselines across multiple reasoning domains. On mathematical reasoning, it achieves the strongest aggregate performance among selection methods and the best result on six of eight benchmarks, while also outperforming competing selection methods on scientific reasoning and code generation. Further analyses show that DeGMatch remains effective across different deconfounding settings and selection budgets, as well as across model families and scales.

Our contributions are as follows:
\begin{itemize}
\item We identify limited target discrimination in target-aware selection on long reasoning trajectories and show that, for gradient-based matching, dominant population-level structure in candidate gradients can obscure finer target-dependent variation.
\item We introduce \textbf{DeGMatch}, which combines target-relevant retrieval with deconfounded, model-specific gradient matching for efficient reasoning specialization.
\item We show that DeGMatch outperforms strong heuristic, target-aware, and reasoning-specific baselines across reasoning domains, and remains effective across deconfounding settings and selection budgets, as well as across model families and scales.
\end{itemize}
\section{Related Work}
\label{sec:related_work}

\subsection{Target-Aware Data Selection}
Target-aware data selection builds on broader data-selection literature, which uses signals such as quality, complexity, or diversity~\citep{chen2024alpagasus,lu2024instag,liu2024deita}, likelihood or learnability~\citep{mindermann2022rholoss,li2024ifd,li2024superfiltering}, semantic representations~\citep{chen2025mig}, and model-dependent utility~\citep{agarwal2025delift}. Target-aware approaches further condition selection on a downstream task or reference set through distribution matching~\citep{xie2023dsir,liu2024tsds}, target similarity~\citep{ivison2025largescale,lee2024instructionmatters}, or optimization-based signals~\citep{xia2024less,feng2024tarot,zhao2025g2is,min2026gist,yang2026targetpaths}. These methods have primarily been studied in instruction-tuning or general data-selection settings, whereas we investigate their transfer to long reasoning trajectories and address limited target discrimination in gradient matching.

\begin{figure*}[t]
    \centering
    \includegraphics[width=\textwidth]{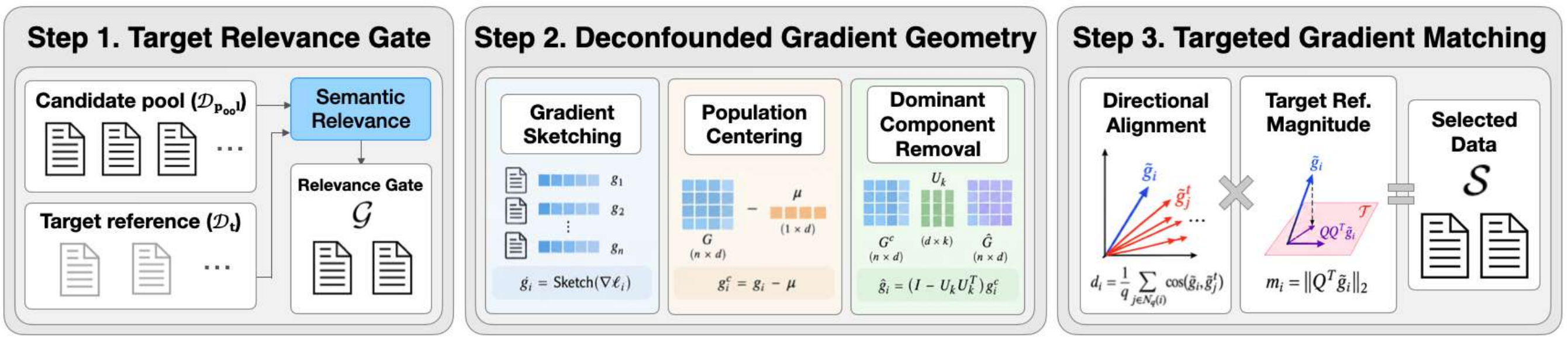}
    \caption{
    Overview of \textbf{DeGMatch}. DeGMatch constructs a target-relevant gate, deconfounds the candidate gradient geometry, and ranks candidates using target-conditioned directional alignment and target-subspace magnitude to select the final training subset.
    }
    \label{fig:method_overview}
    \vspace{-10pt}
\end{figure*}

\subsection{Reasoning-Specific Data Selection}
Reasoning-data curation has explored problem difficulty and trajectory structure~\citep{yang2026select2reason,li2025naturalthoughts,muennighoff-etal-2025-s1,ye2025limoreasoning}, step-level likelihood or uncertainty~\citep{just2025signal,wu2026stards}, and length confounding in likelihood-based selection~\citep{wang2026steplength}. Student-dependent signals include rank--surprisal, token entropy, perturbation-based loss dynamics, learnability, skill coverage, and capability--difficulty alignment~\citep{yang2026informativealignment,li2026unifieddataselectionllm,jin26reasoning,yu2026lark,zhang2026skillaware,yang2026zpd}. Other work adapts the generation or selection of teacher supervision to the student~\citep{he2026studentloop,yun2026cord,cui2026mind,hu2026studentcentric,zhang2026persyn} or uses gradient-related signals for curation and reinforcement learning~\citep{li2026grace,li2026learnalign}. Unlike these approaches, which generally emphasize reasoning-data quality or student suitability, we condition selection on an explicit target capability specified by a small reference set.

\subsection{Gradient-Based Data Selection and Geometry}
Gradient information has been widely used for influence estimation~\citep{koh2017influence,pruthi2020tracin,kwon2024datainf}, subset and coreset selection~\citep{mirzasoleiman2020craig,killamsetty2021glister,killamsetty2021gradmatch,paul2021datadiet}, and target-conditioned data selection~\citep{xia2024less,feng2024tarot,zhao2025g2is,min2026gist}. Prior work also shows that gradient representations can exhibit substantial covariance, low-rank, and spectral structure~\citep{achille2019task2vec,rame2022fishr,wang2023pgrad,fan2026gradientgeometry}, and that normalization or whitening can substantially change their similarity geometry~\citep{ma2024swan,feng2024tarot}. Common-component removal has similarly been used to suppress dominant shared directions in representation spaces~\citep{mu2017allbuttop}. Building on these geometric perspectives, we suppress dominant candidate-population directions before target matching to improve discrimination across targets.

\section{DeGMatch}
\label{sec:method}

DeGMatch selects reasoning trajectories in three stages: it first restricts the candidate pool to target-relevant examples, then removes dominant candidate-population structure from the gradient geometry, and finally ranks candidates through target-conditioned gradient matching. Figure~\ref{fig:method_overview} illustrates the overall procedure.

\subsection{Problem Formulation}
\label{sec:problem_formulation}
Let $\mathcal{D}_{\mathrm{pool}}=\{z_i\}_{i=1}^{N}$ denote a large pool of reasoning trajectories, where $z_i=(x_i,y_i)$ consists of a problem $x_i$ and its reasoning trajectory $y_i$. Let $\mathcal{D}_{\mathrm{target}}=\{a_j\}_{j=1}^{M}$ be a small reference set specifying a target capability, with $a_j=(x_j^t,y_j^t)$. Given a model with parameters $\theta_0$ and a selection budget $B$, our goal is to select a subset $\mathcal{S}\subset\mathcal{D}_{\mathrm{pool}}$, $|\mathcal{S}|=B$, that maximizes the model's post-training performance on the target capability. We refer to this model as the \emph{student} throughout the paper to denote the model being post-trained.

Ideally,
\begin{equation}
    \mathcal{S}^{*}
    =
    \arg\max_{\substack{\mathcal{S}\subset\mathcal{D}_{\mathrm{pool}}\\|\mathcal{S}|=B}}
    \mathcal{U}_{\mathrm{target}}
    \left(
    \operatorname{Train}(\theta_0,\mathcal{S})
    \right),
    \label{eq:selection_objective}
\end{equation}
where $\mathcal{U}_{\mathrm{target}}$ denotes downstream performance on the target capability. Since evaluating this utility for candidate subsets would require repeated post-training, we instead optimize a tractable surrogate. 

\subsection{Target-Relevance Gate}
\label{sec:relevance_gate}

We first restrict gradient-based selection to trajectories whose problems are semantically related to the target capability. Using a frozen embedding model $\phi$, we encode only the problem statements and rank candidates for each target reference by cosine similarity between their normalized embeddings. We construct a relevance gate $\mathcal{G}$ of fixed size $G > B$ by cycling through the target references and adding each reference's highest-ranked unseen candidate until $|\mathcal{G}|=G$. This stage narrows the search space before applying the more expensive student-specific gradient analysis. Embedding similarity is used only to construct $\mathcal{G}$ and does not enter the final DeGMatch score. The resulting gate also defines the candidate population from which we estimate shared gradient structure in the next stage.

\subsection{Deconfounded Gradient Geometry}
\label{sec:deconfounded_gradient}
Within $\mathcal{G}$, we construct model-specific gradient representations for target-conditioned selection. Prior work on representation geometry shows that a shared mean and a small number of dominant principal directions can disproportionately shape pairwise similarity, and that removing these common components can expose more discriminative variation~\citep{mu2017allbuttop}. We find an analogous phenomenon in reasoning gradients: candidate gradients exhibit strongly concentrated population-level structure, while raw target-conditioned rankings remain highly similar across different target references, as detailed in Appendix~\ref{app:deconfounding_analysis}. This motivates removing dominant candidate-population structure before target matching.

\paragraph{Step 1. Gradient sketching.}
Following prior gradient-based data selection and attribution methods that compress high-dimensional per-example gradients for scalable matching~\citep{xia2024less,park2023trak}, we compute the gradient of the token-averaged completion loss with respect to the initial student model and sketch it into a tractable representation:
\begin{equation}
    g_i
    =
    \operatorname{Sketch}
    \left(
    \nabla_{\theta_0}\mathcal{L}(z_i;\theta_0)
    \right),
    \label{eq:gradient_sketch}
\end{equation}
where $\mathcal{L}$ is computed over completion tokens. Target gradients $g_j^t$ are computed using the same student parameters and sketching operator. We use CountSketch~\citep{charikar2002countsketch}; implementation details are provided in Appendix~\ref{app:implementation}.

\paragraph{Step 2. Population centering.}
We first subtract the mean gradient of the candidate population, so that candidate and target gradients are represented relative to the same population mean. We compute
\begin{equation}
    \mu
    =
    \frac{1}{|\mathcal{G}|}
    \sum_{z_i\in\mathcal{G}} g_i,
    \qquad
    \bar g_i=g_i-\mu,
    \qquad
    \bar g_j^t=g_j^t-\mu.
    \label{eq:gradient_centering}
\end{equation}
This centering removes the shared population offset and allows the subsequent principal-component analysis to characterize dominant variation around that baseline. The same candidate-derived mean is applied to target gradients so that candidates and targets share the same centered representation, without using target examples to estimate population statistics.

\paragraph{Step 3. Dominant-component removal.}
Let $U_k$ contain the top $k$ orthonormal principal directions of the centered candidate gradients
$\{\bar g_i:z_i\in\mathcal{G}\}$. We remove this dominant subspace using
\begin{equation}
    P_{\perp}=I-U_kU_k^\top,
\end{equation}
and transform both candidate and target gradients as
\begin{equation}
    \tilde g_i=P_{\perp}\bar g_i,
    \qquad
    \tilde g_j^t=P_{\perp}\bar g_j^t.
    \label{eq:gradient_deconfound}
\end{equation}
Both $\mu$ and $U_k$ are estimated exclusively from the candidate gradients within $\mathcal{G}$, without directly incorporating target gradients into the population statistics. As shown in Appendix~\ref{app:deconfounding_analysis}, raw gradient matching is weakly sensitive to changes in the supplied target references, whereas suppressing this dominant candidate-population subspace substantially increases target sensitivity. This transformation affects only the geometry used for data selection; training uses the original trajectories and standard gradients.

\subsection{Target-Conditioned Gradient Matching}
\label{sec:gradient_matching}

Given the deconfounded gradients, we first compute directional alignment with the target references, following prior gradient-based target matching~\citep{xia2024less,min2026gist}. For candidate $i$, let $\mathcal{N}_q(i)$ denote the $q$ target gradients with the highest cosine similarity, and define
\begin{equation}
    d_i
    =
    \frac{1}{q}
    \sum_{j\in\mathcal{N}_q(i)}
    \cos(\tilde g_i,\tilde g_j^t).
    \label{eq:directional_alignment}
\end{equation}

We additionally weight this alignment by the magnitude of the candidate gradient within the target-gradient subspace. Let
$T=[\tilde g_1^t,\ldots,\tilde g_M^t]$, and let
$Q=\operatorname{orth}(T)$ denote an orthonormal basis for its column span. We define
\begin{equation}
    m_i
    =
    \left\|Q^\top\tilde g_i\right\|_2.
    \label{eq:target_magnitude}
\end{equation}

The two terms are complementary: $d_i$ captures directional agreement with target gradients, while $m_i$ captures the strength of the candidate update within the target-gradient subspace. The final score combines them multiplicatively:
\begin{equation}
    s_i=d_i m_i,
    \qquad
    \mathcal{S}
    =
    \operatorname{TopB}_{z_i\in\mathcal{G}} s_i.
    \label{eq:degmatch_score}
\end{equation}
Implementation details are provided in Appendix~\ref{app:implementation}.

\begin{table*}[t]
\centering
\footnotesize
\setlength{\tabcolsep}{1.0pt}
\renewcommand{\arraystretch}{1.08}

\begin{tabularx}{\textwidth}{
@{}
>{\raggedright\arraybackslash}p{0.14\textwidth}
*{11}{Y}
@{}
}
\toprule
\textbf{Method}
& \textbf{\# Ex.}
& \textbf{Tokens}
& \textbf{AIME24}
& \textbf{AIME25}
& \textbf{AIME26}
& \textbf{AMC23}
& \textbf{HMMT}
& \textbf{MATH500}
& \textbf{GK24}
& \textbf{GK25}
& \textbf{Avg.} \\
\midrule

Base Model
& -- & --
& 1.2 & 1.9 & 0.2 & 12 & 0.2 & 17.3 & 9.4 & 7.2
& 6.18 \\

\midrule

\mbox{OpenR1-Qwen-7B}$^\ddagger$
& 93K & 561.2M
& 47.9 & 34.8 & 35.6 & 84.2 & 24.0 & 92.5 & 72.6 & 78.8
& 58.80 \\

\midrule
\multicolumn{12}{l}{\textit{Heuristic selection}} \\

Random
& 10K & 60.3M
& 39.2 & 30.8 & 34.4 & 77.7 & 18.3 & 89.6 & 70.2 & 80.0
& 55.03 \\

Length
& 10K & 142.3M
& 45.0 & 33.3 & 40.4 & \underline{82.3} & 21.7 & 91.8 & 74.3 & 81.0
& 58.73 \\

MaxPPL
& 10K & 117.2M
& 41.7 & \underline{34.6} & 38.5 & 82.0 & 20.3 & \underline{92.4} & 73.1 & 80.8
& 57.93 \\

\midrule
\multicolumn{12}{l}{\textit{Target-aware data selection}} \\
RDS+
& 10K & 85.2M
& 45.0 & 31.9 & 38.8 & \underline{82.3} & 21.5 & 91.2 & 74.5 & 80.3
& 58.19 \\

LESS
& 10K & 70.6M
& 43.1 & 31.7 & 38.5 & 79.7 & 20.2 & 90.2 & 72.8 & 79.6
& 56.98 \\

GIST
& 10K & 86.1M
& 43.5 & 32.1 & \underline{40.8} & 80.3 & 22.1 & 90.9 & 74.1 & 80.0
& 57.98 \\

TAROT
& 10K & 74.7M
& 39.4 & 34.0 & 40.3 & 78.6 & 21.0 & 91.7 & 70.2 & 81.7
& 57.11 \\

\midrule
\multicolumn{12}{l}{\textit{Reasoning-specific data selection}} \\

RSR
& 10K & 34.9M
& 31.7 & 28.1 & 29.2 & 75.6 & 16.7 & 88.5 & 61.3 & 78.8
& 51.24 \\

TEMP
& 10K & 64.9M
& 42.5 & 30.4 & 35.4 & 81.7 & 19.4 & 91.8 & \underline{76.6} & \underline{82.0}
& 57.48 \\

Select2Reason$^\dagger$
& 19.6K & $\sim$196M
& \underline{45.4} & \textbf{34.8} & 38.1 & 82.0
& \textbf{25.8} & 92.3 & 74.5 & 79.0
& \underline{58.99} \\

\midrule

\textbf{DeGMatch}
& 10K
& 86.9M
& \textbf{47.7}
& 33.8
& \textbf{42.6}
& \textbf{85.6}
& \underline{23.7}
& \textbf{92.8}
& \textbf{76.9}
& \textbf{82.7}
& \textbf{60.73} \\

\bottomrule
\end{tabularx}

\caption{
Main results on mathematical reasoning.
\# Ex. denotes the number of training examples, and Tokens denotes the number of training tokens consumed per epoch.
For target-aware methods, we report the average token count across target-specific training runs, with per-target statistics provided in Appendix~\ref{app:selection_statistics}.
$^\dagger$Select2Reason uses the released Qwen-7B checkpoint, with tokens estimated from reported statistics.
$^\ddagger$OpenR1-Qwen-7B is a full-data reference rather than a budget-matched baseline.
\textbf{Bold} and \underline{underlined} values indicate the best and second-best results among selection methods, respectively.
}
\label{tab:main_math}
\vspace{-10pt}
\end{table*}

\section{Experimental Setup}
\label{sec:experimental_setup}

\subsection{Data and Target References}
\label{sec:data_and_target_references}
We evaluate mathematics, science, and code using OpenR1-Math-220k~\citep{openr1math220k} and the science and CodeForces-CoT portions of Open-R1 Mixture-of-Thoughts~\citep{openr1mot}, yielding 93K, 53K, and 26K trajectories, respectively. Each target uses 50 disjoint reference problems paired with verified DeepSeek-R1 trajectories. Pool and reference-set construction, including contamination filtering, are detailed in Appendices~\ref{app:pool_construction} and~\ref{app:target_references}.

\subsection{Baselines}
\label{sec:baselines}

We compare DeGMatch against heuristic, target-aware, and reasoning-specific data-selection methods. \textbf{Random} uniformly samples from the candidate pool. \textbf{Length} selects the longest trajectories, following prior work showing that response or reasoning-trace length provides a simple but strong signal for post-training data selection~\citep{zhao2024long,yang2026select2reason,guha2026openthoughts,min2026gist}. \textbf{MaxPPL} selects examples with the highest student-model perplexity as a simple proxy for difficulty.

Target-aware baselines include representation-based \textbf{RDS+}~\citep{ivison2025largescale}, gradient-based \textbf{LESS}~\citep{xia2024less} and \textbf{GIST}~\citep{min2026gist}, and optimal-transport-based \textbf{TAROT}~\citep{feng2024tarot}.

Reasoning-specific baselines include \textbf{RSR}~\citep{yang2026informativealignment}, which combines student token rank and surprisal; \textbf{TEMP}~\citep{jin26reasoning}, which uses perturbation-based loss trajectories; and \textbf{Select2Reason}~\citep{yang2026select2reason}, which uses student-relative difficulty and trajectory characteristics.\footnote{For Select2Reason, we evaluate the released \textit{Select2Reason-Qwen-7B} checkpoint (\url{https://huggingface.co/cehao/Select2Reason-Qwen-7B}).} We use matched candidate pools, selection budgets, and downstream training recipes wherever applicable; full implementation and adaptation details are provided in Appendix~\ref{app:baseline_details}.

We additionally report the initial student model and OpenR1-Qwen-7B~\citep{openr1qwen7b} as reference points; the latter is treated as a full-data reference rather than a budget-matched selection baseline.

\subsection{Training Setup}
\label{sec:training_setup}
Our primary student is \textit{open-r1/Qwen2.5-Math-7B-RoPE-300k}~\citep{openr1student}, derived from Qwen2.5-Math-7B~\citep{yang2024qwen25math}. All selected subsets are fine-tuned from the same initialization under an identical training recipe. For the semantic relevance stage of DeGMatch, we use Qwen3-Embedding-8B~\citep{zhang2025qwen3embedding}, and set the relevance-gate size to three times the final selection budget in the main experiments across all domains. We keep all DeGMatch selection hyperparameters fixed across domains without target-specific retuning. Full optimization and implementation details are provided in Appendix~\ref{app:training_details}.

\subsection{Evaluation}
\label{sec:evaluation}

For mathematical reasoning, we build on the evaluation setting used in Select2Reason~\citep{yang2026select2reason} and evaluate on AIME24, AIME25, and AIME26~\citep{maa_aime}, AMC23~\citep{maa_amc}, HMMT Feb.\ 2025~\citep{matharena_hmmt2025}, MATH500~\citep{lightman2024verify}, and Gaokao 2024--2025~\citep{gaokaomath_hf}. For scientific reasoning and code generation, we follow the benchmark families used in OpenThoughts~\citep{guha2026openthoughts}, evaluating on GPQA-Diamond~\citep{rein2024gpqa}, the physics and chemistry portions of JEEBench~\citep{arora2023jeebench}, LiveCodeBench v6~\citep{jain2025livecodebench}, and CodeForces~\citep{openr1codeforces}. Full benchmark provenance, prompting, decoding, verification, and repeated-sampling protocols are provided in Appendix~\ref{app:evaluation_details}.

\section{Main Results}
\label{sec:results}

\subsection{Target-Aware Mathematical Reasoning}
\label{sec:math_results}

Table~\ref{tab:main_math} presents the main results on mathematical reasoning. For target-aware methods, each evaluation target uses its corresponding reference set described in Appendix~\ref{app:target_references}, whereas target-agnostic methods use a single selected subset across benchmarks.

\paragraph{Strong and consistent target-aware specialization.}
DeGMatch achieves the highest aggregate performance among data-selection methods, with an average score of 60.73 across eight benchmarks and the best result on six of eight. It outperforms all budget-matched heuristic, target-aware, and reasoning-specific baselines, as well as the released Select2Reason checkpoint (58.99). DeGMatch also compares favorably with the full-data OpenR1-Qwen-7B reference (58.80), despite selecting only 10K examples from the 93K-example pool, although this full-data comparison is not budget matched.

\begin{table}[t]
\centering
\small
\setlength{\tabcolsep}{4pt}
\renewcommand{\arraystretch}{1.08}

\begin{tabularx}{\columnwidth}{
@{}
>{\raggedright\arraybackslash}p{0.30\columnwidth}
*{4}{Y}
@{}
}
\toprule
\multicolumn{5}{c}{\textbf{Science}} \\
\midrule
\textbf{Method}
& \textbf{\mbox{GPQA-D}}
& \textbf{JEE}
& \textbf{\# Ex.}
& \textbf{Tokens} \\
\midrule

Random
& 46.8 & 45.8 & 5K & 12.2M \\

Length
& \underline{48.2} & 46.9 & 5K & 35.8M \\

RDS+
& 46.4 & \underline{51.6} & 5K & 19.9M \\

TEMP
& 47.5 & 46.6 & 5K & 11.6M \\

\textbf{DeGMatch}
& \textbf{49.0} & \textbf{51.8} & 5K & 22.1M \\

\midrule

Full fine-tuning
& 48.2 & 52.3 & 53K & 131.9M \\

\bottomrule
\noalign{\vskip 6pt}

\toprule
\multicolumn{5}{c}{\textbf{Code}} \\
\midrule
\textbf{Method}
& \textbf{LCB v6}
& \textbf{CF}
& \textbf{\# Ex.}
& \textbf{Tokens} \\
\midrule

Random
& \underline{19.4} & 15.5 & 3K & 39.1M \\

Length
& 16.3 & 13.7 & 3K & 70.7M \\

RDS+
& 18.7 & 15.2 & 3K & 25.8M \\

TEMP
& \textbf{21.4} & \underline{16.2} & 3K & 43.2M \\

\textbf{DeGMatch}
& \textbf{21.4} & \textbf{17.4} & 3K & 41.0M \\

\midrule

Full fine-tuning
& 20.5 & 21.1 & 26K & 337.0M \\

\bottomrule
\end{tabularx}

\caption{
Cross-domain results on scientific reasoning and code generation. Tokens are reported per epoch and averaged across target-specific subsets for target-aware methods. \textbf{Bold} and \underline{underlined} values denote the best and second-best results among selection methods.
}
\label{tab:science_code_results}
\vspace{-10pt}
\end{table}

\paragraph{Existing target-aware signals transfer imperfectly to reasoning.}
Prior target-aware methods such as LESS, RDS+, GIST, and TAROT do not consistently outperform the strong Length baseline in our reasoning setting. LESS, GIST, and TAROT require an auxiliary warm-up or adaptation stage in our implementations, whereas DeGMatch operates directly on gradients from the initial student. These results suggest that target-aware signals developed primarily for instruction tuning do not transfer uniformly to long reasoning trajectories.

\paragraph{Target-specific utility goes beyond length and difficulty.}
Length and MaxPPL remain strong baselines, showing that trajectory length and student difficulty provide useful selection signals. Reasoning-specific methods such as RSR, TEMP, and Select2Reason also perform competitively, yet DeGMatch achieves the strongest aggregate performance while using substantially fewer training tokens than Length and MaxPPL. These results suggest that effective target-specific selection benefits from signals beyond coarse proxies such as trajectory length or difficulty.

\subsection{Generalization Beyond Mathematical Reasoning}
\label{sec:domain_results}

We next examine whether DeGMatch generalizes beyond mathematics by applying the same selection procedure and hyperparameters to scientific reasoning and code generation. We select 5K science trajectories and 3K code trajectories and compare against Random and the strongest applicable baseline from each category in our mathematical experiments: Length, RDS+, and TEMP for heuristic, target-aware, and reasoning-specific selection, respectively.\footnote{Select2Reason is not included because only its released mathematical checkpoint is available; its selected data and selection pipeline are not publicly available for applying the method to science or code.}

\paragraph{Consistent performance across reasoning domains.}
DeGMatch is the strongest or tied-strongest subset-selection method on all four benchmarks without domain-specific retuning. The relative strength of competing signals varies across domains: Length remains competitive on science but degrades on code, RDS+ performs strongly on JEE, and TEMP matches DeGMatch on LiveCodeBench. In contrast, DeGMatch remains consistently strong across both science and code, suggesting that combining target relevance with student-specific gradient utility transfers across different reasoning domains.

\paragraph{Data efficiency persists beyond mathematics.}
Despite using substantially less training data than full fine-tuning, DeGMatch matches or exceeds it on GPQA-D and LiveCodeBench, remains close on JEE, and trails on CodeForces. These results show that target-aware selection remains effective beyond mathematical reasoning with substantially reduced training volume.

\begin{table}[t]
\centering
\footnotesize
\setlength{\tabcolsep}{1.5pt}
\renewcommand{\arraystretch}{1.08}

\begin{tabularx}{\columnwidth}{
@{}
>{\raggedright\arraybackslash}p{0.39\columnwidth}
*{5}{Y}
@{}
}
\toprule
\textbf{Variant}
& \textbf{A24}
& \textbf{A25}
& \textbf{A26}
& \textbf{AMC}
& \textbf{Avg.} \\
\midrule

Random
& 30.4 & 27.3 & 32.3 & 75.5 & 41.38 \\

\midrule
Relevance only
& 32.7 & 30.2 & 32.3 & 79.1 & 43.58 \\

Raw-gradient dir.
& \underline{36.9} & 30.8 & 34.0 & 78.0 & 44.93 \\

Centered dir.
& 33.3 & \textbf{33.1} & \underline{35.6} & 79.1 & 45.28 \\

Deconfounded dir.
& 34.6 & 31.9 & \textbf{36.7} & 80.3 & \underline{45.88} \\

Deconfounded mag.
& 32.5 & \underline{32.5} & 33.3 & \underline{80.5} & 44.70 \\

\midrule
\textbf{DeGMatch}
& \textbf{37.9}
& 31.0
& \underline{35.6}
& \textbf{81.6}
& \textbf{46.53} \\

\bottomrule
\end{tabularx}

\caption{
Component validation for target relevance, candidate-population deconfounding, and the final matching score.
Avg. is the unweighted mean over AIME24, AIME25, AIME26, and AMC23.
\textbf{Bold} and \underline{underlined} values indicate the best and second-best results, respectively.
}
\label{tab:component_ablation}
\vspace{-10pt}
\end{table}

\begin{figure}[t]
    \centering
    \includegraphics[width=0.9\columnwidth]{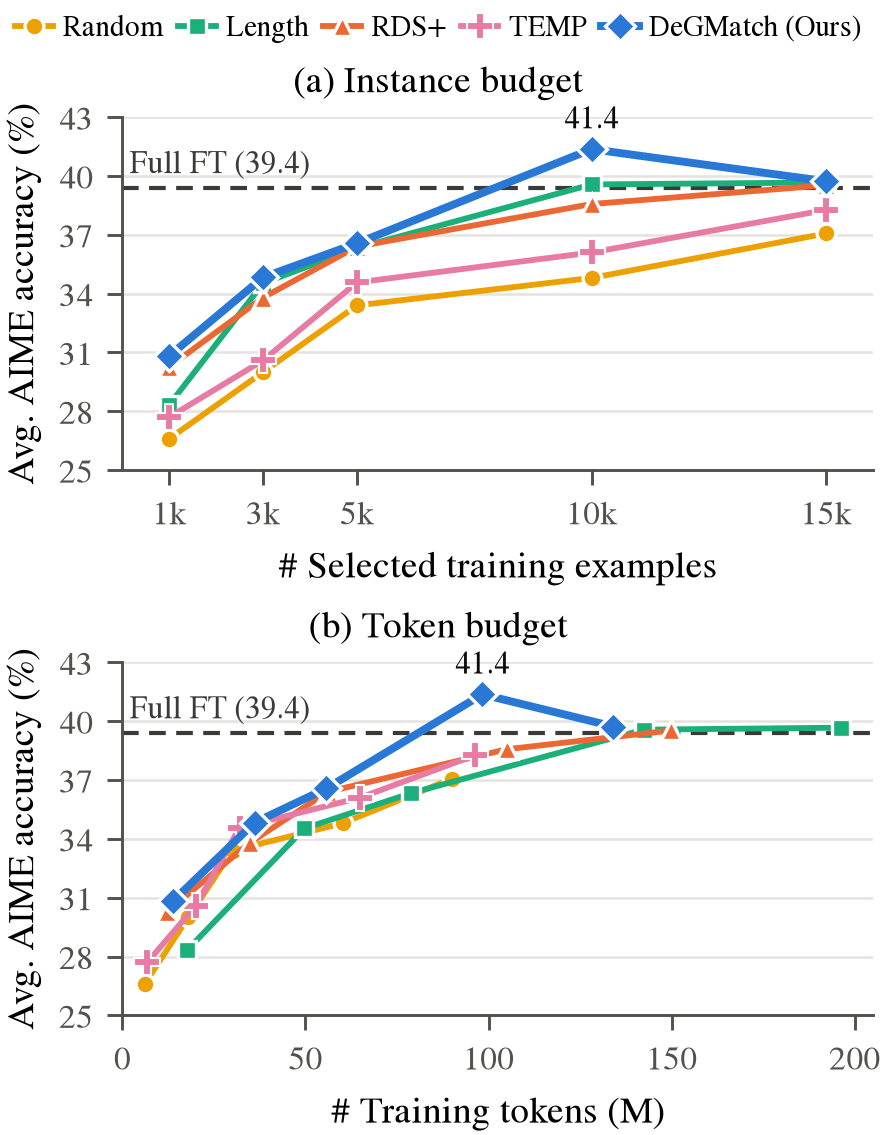}
    \caption{
    Budget scaling by selected examples (top) and training tokens (bottom), measured by the average score on AIME24, AIME25, and AIME26.
    The dashed line denotes full-data training on 93K trajectories (561.2 M tokens).
    Detailed benchmark-level results are provided in Appendix~\ref{app:budget_scaling_details}.
    }
    \label{fig:budget_scaling}
    \vspace{-10pt}
\end{figure}

\section{Analysis}
\label{sec:analysis}

We analyze DeGMatch through component ablations, budget scaling, and cross-model robustness. Unless otherwise specified, analyses use a 3K budget and evaluate on AIME24, AIME25, AIME26, and AMC23 using the same 30K AIME relevance gate; Appendix~\ref{app:gate_size} separately varies the gate size. We compare against Random, Length, RDS+, and TEMP as representative reference, heuristic, target-aware, and reasoning-specific baselines, respectively.

\begin{table}[t]
\centering
\footnotesize
\setlength{\tabcolsep}{2pt}
\renewcommand{\arraystretch}{1.08}

\begin{tabular*}{\columnwidth}{
@{\extracolsep{\fill}}
>{\raggedright\arraybackslash}p{0.22\columnwidth}
*{4}{>{\centering\arraybackslash}p{0.09\columnwidth}}
>{\centering\arraybackslash}p{0.12\columnwidth}
>{\centering\arraybackslash}p{0.13\columnwidth}
@{}
}
\toprule
\textbf{Method}
& \textbf{A24}
& \textbf{A25}
& \textbf{A26}
& \textbf{AMC}
& \textbf{Avg.}
& \textbf{Tokens} \\
\midrule

\multicolumn{7}{@{}l}{\textit{Qwen2.5-3B-Base}} \\
Random   & 7.1 & 5.2 & \underline{6.7} & 39.2 & 14.55 & 18.2M \\
Length   & 7.5 & \underline{7.7} & 5.8 & \textbf{42.2} & \underline{15.80} & 45.8M \\
RDS+     & \underline{7.7} & \textbf{10.4} & 6.2 & 38.7 & 15.75 & 29.1M \\
DeGMatch & \textbf{8.8} & \underline{7.7} & \textbf{7.5} & \underline{40.2} & \textbf{16.05} & 14.7M \\

\midrule

\multicolumn{7}{@{}l}{\textit{Qwen3-1.7B-Base}} \\
Random   & \underline{9.2} & 5.6 & 5.4 & \textbf{36.6} & 14.20 & 17.9M \\
Length   & 7.3 & \underline{8.3} & 5.4 & 30.3 & 12.83 & 49.6M \\
RDS+     & 9 & \textbf{9.2} & \underline{7.1} & 34.5 & \textbf{14.95} & 32.8M \\
DeGMatch & \textbf{9.4} & 6.5 & \textbf{7.2} & \underline{36.3} & \underline{14.85} & 14.8M \\

\midrule

\multicolumn{7}{@{}l}{\textit{Phi-3.5-mini-Instruct}} \\
Random   & 3.1 & 1.2 & 1.7 & 26.2 & 8.05 & 19.4M \\
Length   & 3.1 & 4.0 & \underline{4.2} & 27.0 & 9.58 & 54.2M \\
RDS+     & \underline{6.7} & \underline{4.6} & 4.0 & \underline{28.6} & \underline{10.98} & 34.6M \\
DeGMatch & \textbf{8.6} & \textbf{6.4} & \textbf{5.3} & \textbf{30.0} & \textbf{12.58} & 26.0M \\
\bottomrule
\end{tabular*}

\caption{
Cross-model robustness under a fixed 3K selection budget.
Avg. is the unweighted mean over AIME24/25/26 and AMC23.
Tokens denotes the total number of training tokens in each selected subset.
\textbf{Bold} and \underline{underlined} values denote the best and second-best results among selection methods.
}
\label{tab:cross_model}
\vspace{-10pt}
\end{table}

\subsection{Component Validation}
\label{sec:component_ablation}

We examine the contributions of target relevance, candidate-population deconfounding, and the two terms of the final matching score. Table~\ref{tab:component_ablation} summarizes the results, with detailed variant definitions provided in Appendix~\ref{app:component_details}.

\paragraph{Target relevance provides a useful first-stage signal.}
Relevance-only selection improves over Random, showing that semantic relevance provides a useful signal for identifying target-relevant reasoning trajectories. Its average performance remains below the gradient-based variants and the full method, indicating that relevance alone does not capture student-specific training utility.

\paragraph{Deconfounding improves gradient matching.}
Population centering improves over raw-gradient directional matching on average, and removing the dominant principal directions provides a further gain. This trend is consistent with the diagnostics in Appendix~\ref{app:deconfounding_analysis}, where suppressing dominant candidate-population structure increases the sensitivity of gradient rankings to the supplied target references.

\paragraph{Direction and magnitude provide complementary signals.}
Neither deconfounded directional alignment nor target-subspace magnitude alone matches the full method, while their combination achieves the strongest average performance. Direction captures compatibility with target gradients, whereas magnitude measures the strength of the candidate update within the target-gradient subspace.

\subsection{Data Efficiency Across Selection Budgets}
\label{sec:budget_scaling}
We vary the selection budget from 1K to 15K trajectories and report average performance over AIME24, AIME25, and AIME26 as a function of both the number of selected examples and the resulting training-token volume in Figure~\ref{fig:budget_scaling}.

DeGMatch remains effective across all selection budgets, matching or outperforming the strongest baselines at each scale and achieving its largest advantage at 10K examples, where it also surpasses the full-data reference (41.37 vs. 39.4). The token-based view further shows that these gains are not explained by selecting longer trajectories, as DeGMatch maintains a favorable performance--token trade-off relative to Length and RDS+. Overall, the budget sweep shows that the advantage of DeGMatch is not confined to a particular subset size and that strong performance can be achieved with substantially less training data than full-data training.

\subsection{Robustness Across Student Models}
\label{sec:cross_model}

Finally, we test whether DeGMatch transfers beyond the primary 7B student using three settings: Qwen2.5-3B-Base~\citep{qwen2025qwen25} tests scaling within the same model family, Qwen3-1.7B-Base~\citep{yang2025qwen3} introduces a different Qwen generation at a smaller scale, and Phi-3.5-mini-Instruct~\citep{microsoft2024phi35} provides a cross-family comparison with a different pretraining and instruction-tuning recipe. For each student, we recompute candidate and target gradients and independently select a model-specific subset under the same 3K budget. We exclude TEMP because it requires a parameter-compatible trained endpoint for each student.

Across these settings, DeGMatch remains consistently competitive and achieves the strongest overall performance on two of the three students. Its advantage also does not come from selecting substantially longer trajectories, as it uses fewer training tokens than Length and RDS+ across all three models. These results suggest that the selection principle is not specific to the primary 7B student and transfers across changes in model scale, generation, and family.
\section{Conclusion}
\label{sec:conclusion}

We study \textit{target-aware reasoning data selection}, where a small reference set specifies a target capability to guide the selection of training trajectories. We introduce \textbf{DeGMatch}, which combines target-relevant retrieval with model-specific gradient matching in a deconfounded gradient space. By centering gradients and suppressing dominant candidate-population principal directions, DeGMatch reduces the influence of dominant population-level structure that can obscure target-dependent variation during gradient matching. Across mathematics, science, and code, DeGMatch achieves strong performance against heuristic, target-aware, and reasoning-specific baselines while remaining effective across selection budgets, model families, and scales, supporting deconfounded gradient matching for target-aware reasoning data selection.

\section*{Limitations}
Despite the effectiveness of DeGMatch, several directions remain open. First, our experiments focus on small- to medium-scale student models, and the scalability of student-gradient-based selection to substantially larger models remains to be studied. Second, we consider selection from existing reasoning-trajectory pools; jointly optimizing data generation, evolution, and selection may further improve post-training efficiency. Finally, while our analysis shows that removing dominant shared gradient structure improves target-conditioned matching, understanding what these gradient directions represent in terms of reasoning skills and behaviors remains an open question for future work.

% =========================================================
% Acknowledgments
% =========================================================

% Uncomment for camera-ready version.
% \section*{Acknowledgments}

\bibliography{custom}
\clearpage
\appendix
\section{Experimental Details}
\label{app:experimental_details}

\subsection{Candidate-Pool Construction}
\label{app:pool_construction}

Table~\ref{tab:pool_statistics} summarizes the candidate pools used in our experiments. We retain at most one reasoning trajectory per problem in all domains. Token counts are measured using the student-model tokenizer over the complete training sequence. Problems for which no verified trajectory is obtained within the attempt budget are discarded and replaced by another randomly sampled eligible problem.

For OpenR1-Math-220k \citep{openr1math220k}, we use the \texttt{default} split and retain the first verified DeepSeek-R1 trajectory associated with each problem. For Mixture-of-Thoughts \citep{openr1mot}, we retain one trajectory per problem using a fixed random seed. The resulting science and code pools contain 53,195 and 26,061 unique problems, respectively. No retained example exceeds the 32,768-token training context.

\paragraph{Contamination check.}
We check each candidate pool against its corresponding evaluation benchmarks for both exact overlap and near-duplicate problems. We find no evaluation problems in the final candidate pools under our contamination criteria. This check is performed before data selection, so all selection methods operate on the same contamination-filtered candidate pool.

\paragraph{Analysis metadata.}
For post-hoc analyses of the mathematical candidate pool, we use the subject and source metadata distributed with OpenR1-Math-220k. Subject categories are taken directly from the dataset-provided \texttt{problem\_type} field, which contains Algebra, Geometry, Number Theory, Combinatorics, Logic and Puzzles, Calculus, Inequalities, and Other; we use these labels without remapping. We define the Olympiad indicator as whether the dataset-provided \texttt{source} field equals \texttt{olympiads}. This indicator reflects dataset provenance rather than an inferred measure of difficulty and does not include examples assigned to other contest-related source tags. These metadata are used only for post-hoc analysis of DeGMatch and do not enter its candidate scoring or selection.

\begin{table}[t]
\centering
\small
\setlength{\tabcolsep}{3pt}
\renewcommand{\arraystretch}{1.08}

\begin{tabularx}{\columnwidth}{
@{}
>{\raggedright\arraybackslash}m{0.28\columnwidth}
>{\centering\arraybackslash}m{0.22\columnwidth}
>{\centering\arraybackslash}m{0.20\columnwidth}
>{\centering\arraybackslash}m{0.24\columnwidth}
@{}
}
\toprule
\textbf{Pool} &
\textbf{\# Examples} &
\textbf{Tokens} &
\textbf{Mean Len.} \\
\midrule

Math \citep{openr1math220k}
& 93,733 & 561.2M & 5,986 \\

Science \citep{openr1mot}
& 53,195 & 131.9M & 2,480 \\

Code \citep{openr1mot}
& 26,061 & 337.0M & 12,931 \\

\bottomrule
\end{tabularx}

\caption{Candidate-pool statistics.}
\label{tab:pool_statistics}
\vspace{-10pt}
\end{table}

\subsection{Target-Reference Construction}
\label{app:target_references}

Target references are constructed independently for each target capability from source sets separate from the corresponding evaluation set. For each target, we remove invalid examples and examples belonging to the evaluation set, and then randomly sample 50 reference problems from the remaining pool using a fixed random seed. Each sampled problem is paired with a verified reasoning trajectory generated by DeepSeek-R1 \citep{guo2025deepseekr1}. Table~\ref{tab:target_setup} summarizes the reference source and corresponding evaluation set for each target.

\begin{table*}[t]
\centering
\small
\setlength{\tabcolsep}{4pt}
\renewcommand{\arraystretch}{1.10}

\begin{tabularx}{\textwidth}{
@{}
l
l
>{\raggedright\arraybackslash}X
>{\raggedright\arraybackslash}X
@{}
}
\toprule
\textbf{Domain} &
\textbf{Target} &
\textbf{Reference source} &
\textbf{Evaluation set} \\
\midrule

Math & AIME
& AIME 1983--2023 \citep{gneubig_aime}
& AIME24/25/26 \citep{maa_aime} \\

Math & AMC
& AMC12 2018--2022 \citep{amc12full}
& AMC23 \citep{maa_amc} \\

Math & HMMT
& Omni-MATH HMMT \citep{omnimath_hf}
& HMMT Feb.\ 2025 \citep{matharena_hmmt2025} \\

Math & MATH
& MATH train \citep{hendrycks2021math}
& MATH500 \citep{lightman2024verify} \\

Math & Gaokao
& Gaokao 2023 \citep{gaokao2023en}
& Gaokao 2024/25 \citep{gaokaomath_hf} \\

\midrule

Science & GPQA
& GPQA non-Diamond \citep{rein2024gpqa}
& GPQA-Diamond \citep{rein2024gpqa} \\

Science & JEE
& JEE Main 2023--24 \citep{grafitejee_hf}
& JEEBench \citep{arora2023jeebench} \\

\midrule

Code & LiveCodeBench
& Historical LiveCodeBench \citep{jain2025livecodebench}
& LiveCodeBench v6 \citep{jain2025livecodebench} \\

Code & CodeForces
& CodeForces train \citep{openr1codeforces}
& CodeForces test \citep{openr1codeforces} \\

\bottomrule
\end{tabularx}

\caption{
Target-reference sources and corresponding evaluation sets.
Reference examples specify the target capability and are not directly included as supervised training examples.
}
\label{tab:target_setup}
\vspace{-10pt}
\end{table*}

For mathematical and scientific references, generation uses temperature $0.6$, top-$p=0.95$, and a maximum generation length of 32,768 tokens. We retain the first generated trajectory whose final answer is verified against the reference answer. We allow up to eight generation attempts for AIME, six for HMMT, MATH500, AMC, and Gaokao, and four for GPQA and JEE. Code references are verified by executing the extracted program against available test cases.

\paragraph{AIME.}
We use the \textit{gneubig/aime-1983-2024} collection \citep{gneubig_aime}, which contains historical American Invitational Mathematics Examination problems spanning multiple decades. Reference problems are drawn from competitions through 2023, while evaluation uses later AIME editions.

\paragraph{AMC.}
We use \textit{edev2000/amc12-full} \citep{amc12full}, a collection of historical AMC 12A and 12B competition problems. We use problems from the 2018--2022 competitions as the reference source and evaluate on AMC 2023.

\paragraph{HMMT.}
Historical HMMT problems are obtained from Omni-MATH \citep{omnimath_hf}, a collection of competition-level mathematical problems from multiple sources. We use HMMT problems from this collection as references and evaluate on the February 2025 HMMT benchmark released through MathArena \citep{matharena_hmmt2025}.

\paragraph{MATH.}
We use the training split of the MATH dataset \citep{hendrycks2021math} as the reference source. MATH contains competition-style mathematical problems organized by subject and difficulty level. Evaluation is performed on MATH500 \citep{lightman2024verify}, with exact duplicate problems excluded from the reference pool.

\paragraph{Gaokao.}
We use the English-language Gaokao 2023 mathematics collection \citep{gaokao2023en} as the reference source. Gaokao is China's national college-entrance examination and covers a broad range of high-school mathematics topics. Evaluation uses the 2024 and 2025 English-language mathematics sets from GaoKaoMath \citep{gaokaomath_hf}.

\paragraph{GPQA.}
We use problems from the non-Diamond portion of GPQA \citep{rein2024gpqa} as scientific-reasoning references. GPQA consists of graduate-level multiple-choice questions spanning physics, chemistry, and biology. We evaluate on the more challenging GPQA-Diamond subset.

\paragraph{JEE.}
We construct references from physics and chemistry problems in JEE Main 2023--2024 using the GRAFITe JEE collection \citep{grafitejee_hf}. JEE Main is an Indian engineering entrance examination covering high-school-level physics, chemistry, and mathematics. Evaluation is performed on the physics and chemistry portions of JEEBench \citep{arora2023jeebench}.

\paragraph{LiveCodeBench.}
We use historical problems from LiveCodeBench \citep{jain2025livecodebench} released before September 2024 as references. LiveCodeBench is a continuously updated code-generation benchmark designed to reduce contamination from previously released programming problems. Evaluation uses version 6 problems released between January and April 2025.

\paragraph{CodeForces.}
We use the training split of the Open-R1 CodeForces collection \citep{openr1codeforces} as the reference source and its disjoint test split for evaluation. Reference problems are restricted to non-interactive problems with official test cases and are excluded if they already occur in the Mixture-of-Thoughts code candidate pool.

\paragraph{Relevance gate.}
We use Qwen3-Embedding-8B \citep{zhang2025qwen3embedding} as the frozen embedding model. Only the problem statement is embedded; reasoning trajectories and auxiliary instructions are excluded. Inputs are truncated to 2,048 tokens, left padded, and represented by the final-token hidden state followed by $\ell_2$ normalization.

For each target reference, we independently rank the complete candidate pool by cosine similarity. Starting from retrieval depth $k=1$, we cycle through target references in a fixed order and consider the $k$-th nearest candidate of each reference. A candidate is added only if it has not already been included in the gate; duplicates are skipped. We then advance to the next retrieval depth and repeat until the desired gate size is reached. We set the gate size to three times the final selection budget, corresponding to 30K candidates for mathematics, 15K for science, and 9K for code. Embedding similarity is used only to construct the gate and is discarded before gradient-based reranking.

\subsection{DeGMatch Implementation}
\label{app:implementation}

\begin{table*}[t]
\centering
\small
\setlength{\tabcolsep}{4pt}
\renewcommand{\arraystretch}{1.08}

\begingroup
\renewcommand{\tabularxcolumn}[1]{m{#1}}
\begin{tabularx}{\textwidth}{
@{}
>{\raggedright\arraybackslash}m{0.12\textwidth}
>{\raggedright\arraybackslash}m{0.18\textwidth}
>{\raggedright\arraybackslash}X
>{\raggedright\arraybackslash}m{0.25\textwidth}
@{}
}
\toprule
\textbf{Method}
& \textbf{Implementation source}
& \textbf{Preserved selection rule}
& \textbf{Adaptation to our setting} \\
\midrule

RDS+
& Published procedure
& Position-weighted student hidden-state representations, cosine similarity, and round-robin target retrieval
& Initial student representations computed over complete sequences up to 32K tokens \\
\midrule

LESS
& Official pipeline
& Adam-preconditioned gradients across warmup checkpoints with learning-rate-weighted target alignment
& Common LoRA warmup; per-example gradients computed over complete sequences up to 32K tokens \\
\midrule

GIST
& Official pipeline
& Target-gradient Gram decomposition, 95\% spectral-mass subspace, and maximum target-reference cosine
& Common LoRA warmup; gradients computed over complete sequences up to 32K tokens \\
\midrule

TAROT
& Published method description
& Multi-checkpoint gradient features, whitening, and fixed-budget OT-based selection
& Common LoRA warmup; gradients computed over complete sequences up to 32K tokens \\
\midrule

RSR
& Official code
& Rank--surprisal ratio from student-model response-token probabilities
& Global fixed-budget ranking rather than within-problem trajectory selection \\
\midrule

TEMP
& Official code + adaptation
& Difficulty filtering, perturbation-based loss trajectories, clustering, and learnability
& OpenR1-Distill-7B as directional endpoint; source allocation follows the published equations \\
\midrule

Select2Reason
& Released checkpoint
& --
& External reference; starts from a RoPE-extended Qwen2.5-Math-7B-Instruct model rather than our base student \\

\bottomrule
\end{tabularx}
\endgroup

\caption{
Implementation sources and adaptations for the principal comparison methods. For methods producing selected subsets in our pipeline, we preserve the published selection rule while using the same candidate pool, final selection budget, and downstream fine-tuning recipe as DeGMatch.
}
\label{tab:baseline_reproduction}
\vspace{-10pt}
\end{table*}

\paragraph{Gradient sketching.}
Directly storing a full-parameter gradient for every trajectory is impractical. Prior scalable gradient-based methods therefore construct low-dimensional gradient features before similarity or influence estimation; for example, LESS uses an 8,192-dimensional Rademacher random projection~\citep{xia2024less}, following the random-projection approach used in TRAK~\citep{park2023trak}. We follow this general practice but sketch the full-parameter gradients using CountSketch~\citep{charikar2002countsketch}. Specifically, each per-example gradient is projected to 16,384 dimensions using a fixed sketch shared across all candidate and target examples. Prompt tokens are masked, and the loss is averaged over completion tokens.

The sketch in Step 1 is a fixed computational compression applied uniformly to candidates and targets, whereas the projection in Step 3 is the candidate-dependent transformation that deliberately removes dominant population structure before matching.

\paragraph{Deconfounding.}
The gradient mean and principal directions are estimated only from examples inside the relevance gate. We subtract the gate-gradient mean and remove the top 50 principal directions. The same mean and projection are subsequently applied to target gradients. The target references therefore do not determine which population-level directions are removed. 
Let
\[
G=
\begin{bmatrix}
\bar g_1^\top\\
\vdots\\
\bar g_{|\mathcal G|}^\top
\end{bmatrix}
\in\mathbb{R}^{|\mathcal G|\times d}
\]
denote the matrix of centered candidate gradients. We compute $G=L\Sigma R^\top$ and define the dominant gradient-space basis as $U_k=R_{:,1:k}$, i.e., the top $k$ right singular vectors of the centered candidate-gradient matrix.

\paragraph{Target matching.}
For each transformed candidate gradient, we compute cosine similarity to all transformed target gradients and define directional alignment as the mean similarity to the five most aligned references; thus, $q=5$ in Eq.~\ref{eq:directional_alignment} for all experiments. To compute target-subspace magnitude, we form
\[
T=[\tilde g_1^t,\ldots,\tilde g_M^t]
\]
and obtain an orthonormal basis $Q$ for its column space. We then compute $m_i=\|Q^\top\tilde g_i\|_2$ and rank candidates using the unclamped product $s_i=d_i m_i$. The top-$B$ candidates within the relevance gate are selected for training.

\subsection{Baseline Implementation Details}
\label{app:baseline_details}

We implement all baselines under a controlled comparison while preserving their published selection rules as closely as possible. Unless otherwise noted, candidate scoring uses the same initial student model $\theta_0$, examples are formatted using the same chat template as downstream training, and student-model scores are computed over completion tokens only with sequences retained up to 32,768 tokens. All selected subsets are subsequently fine-tuned from the same initial student checkpoint using the common training recipe in Appendix~\ref{app:training_details}.

Table~\ref{tab:baseline_reproduction} summarizes the implementation source, preserved selection rule, and main adaptation used for each principal baseline.

\subsubsection{Heuristic Baselines}
\paragraph{Random.}
We uniformly sample examples from a fixed random permutation of the candidate pool.

\paragraph{Length.}
We rank trajectories by the number of student-tokenizer tokens in the complete training sequence and select the longest examples.

\paragraph{MaxPPL.}
We compute the mean negative log-likelihood over completion tokens under the initial student model and select examples with the highest values.

\subsubsection{Target-Aware Baselines}
\paragraph{RDS+.}
We reimplement RDS+~\citep{ivison2025largescale} following the published procedure. The representation model is the same initial student checkpoint used for downstream training, Qwen2.5-Math-7B-RoPE-300k, without additional warmup or fine-tuning. Each candidate is formatted exactly as its training sequence, including the system prompt, problem, and complete reasoning trajectory in the student chat template; target references are formatted analogously using the reference problem and its verified reasoning trajectory.

We run a single bf16 forward pass over the full sequence, up to 32,768 tokens, and extract final-layer hidden states $h_1,\ldots,h_L$. Following RDS+, the sequence representation is the position-weighted mean
\begin{equation}
e(z)
=
\sum_{i=1}^{L} w_i h_i,
\qquad
w_i
=
\frac{i}{\sum_{j=1}^{L}j},
\end{equation}
computed in fp32 and subsequently $\ell_2$-normalized. Candidate--reference similarity is measured by cosine similarity. We aggregate the resulting rankings using round-robin retrieval: cycling through target references in a fixed order, each reference contributes its highest-ranked candidate not already selected until the budget is reached. This construction is deterministic, and selections at smaller budgets are prefixes of those at larger budgets.

\paragraph{Shared warmup and gradient extraction.}
LESS~\citep{xia2024less}, GIST~\citep{min2026gist}, and TAROT~\citep{feng2024tarot} require gradient features obtained after an auxiliary warmup or adaptation stage. To avoid introducing variation from independently trained warmup models, we use a shared LoRA warmup for these three baselines.

Starting from the initial student, we train a LoRA~\citep{hu2022lora} adapter with rank 128, $\alpha=512$, and dropout 0.1 on the query, key, value, and output projections, yielding 80.7M trainable parameters. The adapter is trained on a uniformly sampled 5\% of the mathematical candidate pool (4,683 examples, seed 42) for four epochs using AdamW with learning rate $2\times10^{-5}$, $\beta=(0.9,0.999)$, $\epsilon=10^{-8}$, no weight decay, a linear schedule with 3\% warmup, effective batch size 32, and bf16 precision. We save the adapter and optimizer state after every epoch, yielding checkpoints $\theta_1,\ldots,\theta_4$.

For every candidate and target-reference example, we compute the per-example gradient of the mean token-level cross-entropy over completion tokens with respect to the LoRA parameters, using batch size 1 and disabling dropout. Because the original instruction-tuning implementations operate on substantially shorter inputs, we retain complete sequences up to 32,768 tokens to accommodate long reasoning trajectories and match our downstream training context. Target references are represented as problem--reasoning-trajectory pairs.

Whenever a baseline requires random projection, we use a fixed Rademacher matrix
\begin{equation}
    \Pi \in \{-1,+1\}^{8192\times d},
\end{equation}
shared across checkpoints, candidates, and target references.

\paragraph{LESS.}
Following LESS~\citep{xia2024less}, candidate examples are represented using Adam-preconditioned gradients. For a candidate gradient $g(z;\theta_i)$ at checkpoint $\theta_i$, we compute
\begin{equation}
\Gamma(z;\theta_i)
=
\frac{
    \beta_1 m_i + (1-\beta_1)g(z;\theta_i)
}{
    \sqrt{
        \beta_2 v_i + (1-\beta_2)g(z;\theta_i)^2 + \epsilon
    }
},
\end{equation}
where $(m_i,v_i)$ are the first- and second-moment optimizer states stored at checkpoint $\theta_i$. We project these features to 8,192 dimensions using $\Pi$.

For each checkpoint, target references use plain gradients with the same projection and are averaged within the target capability to obtain $\bar{g}^{\,t}_i$. We score each candidate using
\begin{equation}
s_{\mathrm{LESS}}(z)
=
\sum_{i=1}^{4}
\bar{\eta}_i
\cos\!\left(
    \Pi\Gamma(z;\theta_i),
    \bar{g}^{\,t}_i
\right),
\end{equation}
where $\bar{\eta}_i$ is the average learning rate during warmup epoch $i$, and select the highest-scoring examples.

\paragraph{GIST.}
Following GIST~\citep{min2026gist}, we use the first warmup checkpoint $\theta_1$ and plain LoRA gradients without Adam preconditioning or random projection. Let
\begin{equation}
    G \in \mathbb{R}^{M\times d}
\end{equation}
stack the gradients of the $M$ target-reference examples. We eigendecompose the uncentered target-gradient Gram matrix,
\begin{equation}
    GG^\top
    =
    V\Lambda V^\top,
\end{equation}
and retain the smallest rank $r$ explaining at least 95\% of the total eigenvalue mass; across our mathematical targets, $r$ ranges from 34 to 37.

Candidate gradients are represented in the corresponding target-gradient subspace as
\begin{equation}
U_r^\top g
=
\Lambda_r^{-1/2}V_r^\top(Gg),
\end{equation}
which we compute directly from candidate--reference gradient inner products. Target references are represented by the corresponding coordinates $V_r\Lambda_r^{1/2}$. Each candidate is scored by its maximum cosine similarity to an individual target reference in this subspace, and we select the highest-scoring examples.

\paragraph{TAROT.}
For TAROT~\citep{feng2024tarot}, we follow the published method description; the released implementation does not provide an LLM-specific training pipeline. The influence feature for an example is the sum of its projected plain gradients across the four warmup checkpoints,
\begin{equation}
\phi(z)
=
\sum_{i=1}^{4}
\Pi g(z;\theta_i).
\end{equation}

We mean-center candidate and reference features and apply Cholesky whitening using their joint covariance, regularized by $10^{-5}I$, followed by $\ell_2$ normalization. Pairwise cost is the Euclidean distance between the resulting whitened features,
\begin{equation}
c(z,z')
=
\left\|
\hat{\phi}(z)-\hat{\phi}(z')
\right\|_2.
\end{equation}

We use TAROT's fixed-size selection procedure with the same final budget as DeGMatch. Candidates are added layer by layer, where layer $\ell$ contains the $\ell$-th nearest candidate for each target reference. If adding a complete layer would exceed the budget, candidates in the final layer are ordered by their dual potentials under the optimal-transport problem between the provisional selected set and the reference set.

We solve the OT problem using entropic Sinkhorn iterations implemented with GeomLoss~\citep{feydy2019interpolating}, with blur 0.01 ($\epsilon=10^{-4}$), $\epsilon$-scaling 0.5, no debiasing, and uniform masses. Across our targets, the budget is reached after 218--250 layers, so the OT-based tie breaking affects fewer than 0.5\% of selected examples. We do not use OT-based reweighting or automatic selection-size estimation.

\subsubsection{Reasoning-Specific Baselines}

Several reasoning-specific methods were originally evaluated under settings that differ from our one-trajectory-per-problem candidate pool. Where necessary, we preserve their trajectory-level scoring or selection rule but adapt the final selection to a global fixed budget over our candidate pool. We treat such implementations as faithful adaptations under a controlled comparison.

\paragraph{RSR.}
We faithfully adapt Rank--Surprisal Ratio (RSR)~\citep{yang2026informativealignment} using the authors' released implementation as reference. For each completion token $t$, let $p_t$ denote the probability assigned by the initial student to the observed token and let $r_t$ denote its rank under the next-token distribution. Following the published implementation, ranks are truncated at 100, and the trajectory score is
\begin{equation}
    s_{\mathrm{RSR}}(z)
    =
    \frac{
        \sum_{t\in y}\min(r_t,100)
    }{
        \sum_{t\in y}-\log p_t
    }.
\end{equation}
Lower scores are preferred.

The original setting selects among multiple teacher trajectories associated with the same problem. Because our pool contains at most one trajectory per problem, we preserve the trajectory-level score but rank all candidates globally and select the lowest-scoring $B$ examples. Our implementation follows the released definition of token rank, rank truncation, and response-token boundaries. We use the official \texttt{RankSurprisalRatio} repository (\url{https://github.com/UmeanNever/RankSurprisalRatio}) at commit \texttt{59a7c4c} as the implementation reference.

\paragraph{TEMP.}
We faithfully adapt TEMP~\citep{jin26reasoning} using the authors' official implementation. TEMP estimates difficulty, diversity, and learnability from losses measured under perturbed versions of the student model. We use the initial student as $\theta_0$ and \textit{open-r1/OpenR1-Distill-7B}~\citep{openr1distill7b} as the endpoint model $\theta_f$ defining the directional perturbation axis.

We first perform the random-perturbation stage used to identify difficult examples. Gaussian perturbations are cumulatively applied to all non-embedding parameters of $\theta_0$ at eight increasing perturbation levels. On a fixed 2K calibration subset, we select the perturbation checkpoint whose mean loss ratio relative to $\theta_0$, measured over the first 100 completion tokens, falls in the prescribed range of $[2,3]$ and is closest to 2.5. For each source in the mathematical pool, we then apply $k$-means with $k=2$ to these losses and retain the higher-loss cluster. This leaves 42,862 of the original 93,733 mathematical candidates.

For the retained candidates, we construct loss trajectories along the direction from $\theta_0$ to $\theta_f$. Let
\begin{equation}
    e
    =
    \frac{\theta_f-\theta_0}
         {\|\theta_f-\theta_0\|_2}.
\end{equation}
Following the released perturbation procedure, we generate eight directional checkpoints with tensor-wise seeded perturbations along $e$ and record the mean loss over the first 1,000 completion tokens at each checkpoint. Each candidate is therefore represented by an eight-dimensional loss trajectory.

We then follow TEMP's source-aware selection procedure. Source-level budgets are allocated from the published difficulty statistics, loss trajectories are clustered independently within each source using FAISS $k$-means, and candidates within each cluster are prioritized by their loss reduction from the first to the final directional checkpoint. The number of clusters within each source is set to approximately one quarter of its allocated budget, following the released selection script.

The released implementation computes the source allocation through a routine that assumes increasing loss along the directional checkpoints. This assumption does not hold for our reasoning-tuned endpoint model, for which loss generally decreases toward $\theta_f$. We therefore compute source allocations directly from the published equations while preserving the remaining perturbation, clustering, and within-cluster selection procedure. The resulting mathematical subset contains 10,000 trajectories and 64.9M training tokens. We use the official \texttt{TEMP-ICML} repository (\url{https://github.com/BigML-CS-UCLA/TEMP-ICML}) at commit \texttt{b2a08ec} as the implementation reference. We apply the same TEMP procedure to the mathematical, scientific, and code candidate pools wherever the required endpoint-model scoring and downstream evaluation are available. Accordingly, we report TEMP on all mathematical, science and code benchmarks.

\begin{table*}[!tp]
\centering
\scriptsize
\setlength{\tabcolsep}{2.7pt}
\renewcommand{\arraystretch}{1.10}

\begin{tabularx}{\textwidth}{
@{}
l
l
c
c
c
>{\raggedright\arraybackslash}X
c
>{\raggedright\arraybackslash}p{0.14\textwidth}
@{}
}
\toprule
\textbf{Method}
& \textbf{Setting}
& \textbf{Target}
& \textbf{Reason.}
& \textbf{Student}
& \textbf{Primary signal}
& \textbf{Grad.}
& \textbf{Preparation} \\

\midrule
AlpaGasus \citep{chen2024alpagasus}
& Instruction & No & No & No
& LLM-estimated quality
& No
& LLM scoring \\

InsTag \citep{lu2024instag}
& Instruction & No & No & No
& Instruction complexity / diversity
& No
& Tagging model \\

DEITA \citep{liu2024deita}
& Instruction & No & No & No
& Quality, complexity, diversity
& No
& Scoring models \\

RHO-LOSS \citep{mindermann2022rholoss}
& General & No & No & Yes
& Reducible loss / learnability
& No
& Reference-model fit \\

IFD \citep{li2024ifd}
& Instruction & No & No & Yes
& Instruction-following difficulty
& No
& Model scoring \\

SuperFiltering \citep{li2024superfiltering}
& Instruction & No & No & No
& Surrogate-model difficulty
& No
& Surrogate scoring \\

GraNd \citep{paul2021datadiet}
& General & No & No & Yes
& Gradient norm
& Yes
& Gradient pass \\

CRAIG \citep{mirzasoleiman2020craig}
& General & No & No & Yes
& Gradient-space coverage
& Yes
& Gradient pass \\

GLISTER \citep{killamsetty2021glister}
& General & Yes & No & Yes
& Validation-gradient utility
& Yes
& Gradient estimation \\

GradMatch \citep{killamsetty2021gradmatch}
& General & No & No & Yes
& Gradient matching
& Yes
& Gradient estimation \\

MIG \citep{chen2025mig}
& Instruction & No & No & No
& Representation geometry
& No
& Representation extraction \\

DELIFT \citep{agarwal2025delift}
& Instruction & No & No & Yes
& Estimated downstream utility
& No
& Utility modeling \\

\midrule

DSIR \citep{xie2023dsir}
& General & Yes & No & No
& Distribution matching
& No
& Density estimation \\

TSDS \citep{liu2024tsds}
& General & Yes & No & No
& Target distribution matching + diversity
& No
& Representation / density estimation \\

RDS+ \citep{ivison2025largescale}
& Instruction & Yes & No & Yes
& Representation-based target relevance
& No
& None \\

LESS \citep{xia2024less}
& Instruction & Yes & No & Yes
& Optimizer-aware source--target alignment
& Yes
& Warmup \\

G2IS \citep{zhao2025g2is}
& Instruction & Yes & No & Yes
& Gradient graph structure
& Yes
& Gradient features \\

GIST \citep{min2026gist}
& Instruction & Yes & No & Yes
& Target-gradient subspace
& Yes
& Warmup \\

TAROT \citep{feng2024tarot}
& General & Yes & No & Yes
& Whitened influence geometry + OT
& Yes
& Influence features \\

\midrule
Select2Reason \citep{yang2026select2reason}
& Reasoning & No & Yes & Yes
& Difficulty + trajectory characteristics
& No
& Difficulty estimation \\

NaturalThoughts \citep{li2025naturalthoughts}
& Reasoning & No & Yes & No
& Difficulty + reasoning diversity
& No
& None \\

RSR / Informative Alignment \citep{yang2026informativealignment}
& Reasoning & No & Yes & Yes
& Token rank + surprisal
& No
& Student scoring \\

TEMP \citep{jin26reasoning}
& Reasoning & No & Yes & Yes
& Perturbation-based loss trajectories
& No
& Endpoint model + clustering \\

HES \citep{li2026unifieddataselectionllm}
& Reasoning & No & Yes & Yes
& High-entropy token sum
& No
& Student scoring \\

LARK \citep{yu2026lark}
& Reasoning & No & Yes & Yes
& Learnability + coverage
& No
& Learnability estimation \\

Signal in Steps \citep{just2025signal}
& Reasoning & No & Yes & Yes
& Step-level supervision signal
& No
& Step scoring \\

Star-DS \citep{wu2026stards}
& Reasoning / RL & No & Yes & Yes
& Step-level uncertainty
& No
& Policy scoring \\

ASLEC \citep{wang2026steplength}
& Reasoning & No & Yes & Yes
& Step-length-corrected log probability
& No
& Token scoring / bias correction \\

Skill-Aware Selection \citep{zhang2026skillaware}
& Reasoning & No & Yes & Yes
& Skill coverage / student capability
& No
& Skill estimation \\

GRACE \citep{li2026grace}
& Reasoning & No & Yes & Yes
& Step-level gradient-proxy alignment
& Proxy
& Warmup + forward-only scoring \\

LearnAlign \citep{li2026learnalign}
& Reasoning / RL & No & Yes & Yes
& Learnability + gradient alignment
& Yes
& Learnability / gradient estimation \\

ZPD \citep{yang2026zpd}
& General & No & No & Yes
& Capability--difficulty alignment
& No
& Difficulty / capability estimation \\

\textbf{DeGMatch}
& \textbf{Reasoning}
& \textbf{Yes}
& \textbf{Yes}
& \textbf{Yes}
& \textbf{Target relevance + deconfounded student-gradient matching}
& \textbf{Yes}
& \textbf{No warmup; gradient sketches + PCA} \\

\bottomrule
\end{tabularx}

\caption{
Extended comparison of representative data-selection methods.
\emph{Target} indicates explicit conditioning on an external target or reference set;
\emph{Reason.} indicates that the method is designed specifically for reasoning data;
\emph{Student} indicates use of student-model signals rather than surrogate-only scoring.
\emph{Grad.} distinguishes gradient-based signals from forward-only gradient proxies.
The final column summarizes auxiliary fitting, scoring, or feature construction required for selection rather than quantitative compute.
}
\label{tab:selection_landscape}
\vspace{-10pt}
\end{table*}

\paragraph{Select2Reason.}
For Select2Reason~\citep{yang2026select2reason}, the data-selection code and selected training data are not publicly available. We therefore evaluate the authors' released \textit{cehao/Select2Reason-Qwen-7B} checkpoint rather than reproducing its data-selection and fine-tuning pipeline. The released checkpoint is initialized from a RoPE-extended Qwen2.5-Math-7B-Instruct model, whereas DeGMatch and our matched implementations start from Qwen2.5-Math-7B-RoPE-300k, which is derived from the base Qwen2.5-Math-7B model and has not undergone instruction or reasoning post-training. Select2Reason therefore begins from a more specialized mathematical instruction-tuned initialization in addition to using a larger selected training set. We treat it as an external reasoning-selection reference rather than a matched implementation under our common training recipe.

\subsection{Extended Comparison of Data-Selection Methods}
Table~\ref{tab:selection_landscape} summarizes representative data-selection methods along dimensions relevant to our setting, including target conditioning, reasoning-specific design, student dependence, selection signal, use of gradients, and auxiliary fitting. The comparison provides additional context for the reasoning-specific and target-aware selection methods discussed in Section~\ref{sec:related_work}.

\begin{table*}[t]
\centering
\small
\setlength{\tabcolsep}{4pt}
\renewcommand{\arraystretch}{1.06}

\begin{tabularx}{\textwidth}{
@{}
>{\raggedright\arraybackslash}p{0.48\textwidth}
>{\centering\arraybackslash}p{0.11\textwidth}
>{\centering\arraybackslash}p{0.11\textwidth}
>{\raggedright\arraybackslash}X
@{}
}
\toprule
\textbf{Benchmark}
& \textbf{\# Problems}
& \textbf{$K$ (Avg@$K$)}
& \textbf{Scoring} \\
\midrule

\mbox{AIME24/25/26 \citep{maa_aime}}
& 30 / 30 / 30 & 16 & final-answer equivalence \\

AMC23 \citep{maa_amc}
& 40 & 16 & final-answer equivalence \\

HMMT Feb.\ 2025 \citep{matharena_hmmt2025}
& 30 & 16 & symbolic answer verification \\

MATH500 \citep{lightman2024verify}
& 500 & 4 & \texttt{math-verify} \\

Gaokao 2024/25 \citep{gaokaomath_hf}
& 26 / 26 & 16 & \texttt{math-verify} \\

GPQA-Diamond \citep{rein2024gpqa}
& 198 & 8 & multiple-choice accuracy \\

JEEBench \citep{arora2023jeebench}
& 279 & 4 & official benchmark scoring \\

LiveCodeBench v6 \citep{jain2025livecodebench}
& 182 & 4 & execution-based correctness \\

CodeForces \citep{openr1codeforces}
& 422 & 4 & official-test execution \\

\bottomrule
\end{tabularx}

\caption{
Evaluation protocols.
We report Avg@$K$, the average correctness across $K$ sampled generations per problem.
}
\label{tab:evaluation_details}
\vspace{-10pt}
\end{table*}

\paragraph{Scope of the empirical baseline comparison.}
Our main experiments are not intended to exhaustively reproduce every data-curation method listed in Table~\ref{tab:selection_landscape}. Instead, we select representative baselines that span the principal selection signals most directly comparable to our fixed-pool supervised fine-tuning setting. The target-aware baselines cover representation-based retrieval (RDS+), optimizer-aware gradient alignment (LESS), target-gradient subspace matching (GIST), and optimal-transport-based influence matching (TAROT). The reasoning-specific comparisons cover student-relative token difficulty (RSR), perturbation-based loss dynamics (TEMP), and Select2Reason as a recent reasoning-specific selection reference. Other methods in Table~\ref{tab:selection_landscape} provide broader context but may rely on different supervision, training objectives, data-construction assumptions, or closely related selection signals. We therefore use the extended table to situate DeGMatch within the broader literature while focusing the controlled experiments on representative methods that admit the closest comparison under our candidate pools and training setup.

\subsection{Training Details}
\label{app:training_details}
Unless otherwise specified, subset experiments start from \texttt{open-r1/Qwen2.5-Math-7B-RoPE-300k} \citep{openr1student}, derived from Qwen2.5-Math-7B \citep{yang2024qwen25math}. We perform full-parameter supervised fine-tuning using FSDP with bfloat16 precision and gradient checkpointing. We use AdamW with $\beta_1=0.9$, $\beta_2=0.95$, weight decay $10^{-4}$, a peak learning rate of $3\times10^{-5}$, cosine decay, and 5\% warmup. The global batch size is 32, and each subset is trained for five epochs with a maximum sequence length of 32,768 tokens. We apply completion-only loss to assistant-response tokens and do not use sequence packing.

The training sequence uses the Qwen chat format with the system message
\texttt{"You are Qwen, created by Alibaba Cloud. You are a helpful assistant."}
All subset experiments use training seed 42 and evaluate the final checkpoint. For full-data fine-tuning, we follow the OpenR1 training recipe, using a learning rate of $5\times10^{-5}$, a global batch size of 128, and three training epochs. This recipe is used only for the full-data reference.

\subsection{Evaluation Details}
\label{app:evaluation_details}

We use a maximum model and generation length of 32,768 tokens. Evaluation is performed with vLLM using the model's chat template. Table~\ref{tab:evaluation_details} summarizes benchmark size and repeated-sampling settings.

The initial student is a pretrained Qwen2.5-Math-7B base checkpoint whose RoPE base frequency was increased to 300K to support a 32K context window; it has not undergone instruction or reasoning post-training. Under this zero-shot evaluation setup, 21.4\% of its generations reach the 32K-token limit, and outputs frequently continue beyond an attempted answer with new problems or repetitive text. We therefore report its score as a reference for the initial student checkpoint rather than as a competitive reasoning baseline.

\begin{table*}[t]
\centering
\small
\setlength{\tabcolsep}{4pt}
\renewcommand{\arraystretch}{1.08}

\begin{tabularx}{\textwidth}{
@{}
>{\raggedright\arraybackslash}p{0.10\textwidth}
>{\raggedright\arraybackslash}p{0.25\textwidth}
>{\raggedright\arraybackslash}X
>{\centering\arraybackslash}p{0.15\textwidth}
@{}
}
\toprule
\textbf{Method}
& \textbf{Auxiliary fitting}
& \textbf{Reusable candidate representation}
& \textbf{Approx.\ GPU-h} \\
\midrule

RDS+
& None
& Student hidden states (full-sequence forward)
& $\sim$12 \\

LESS
& Four-epoch LoRA warm-up
& Optimizer-aware LoRA gradients (4 checkpoints)
& $\sim$200 \\

GIST
& LoRA warm-up
& LoRA gradients at warm-up checkpoint
& $\sim$57 \\

TAROT
& Four-epoch LoRA warm-up
& Projected LoRA gradients (4 checkpoints)
& $\sim$200 \\

DeGMatch
& None
& Full-parameter CountSketch gradients at initialization
& $\sim58.7$ \\

\bottomrule
\end{tabularx}

\caption{
Approximate one-time selection-feature construction cost for the mathematical candidate pool under our 32K-context implementations.
Candidate features can be reused across target capabilities.
The DeGMatch value includes the Qwen3-Embedding-8B problem embedding of the pool used for the relevance gate ($\sim$0.3 GPU-h), the full-pool gradient pass, and one 50-example target-gradient computation.
LESS, GIST, and TAROT use LoRA gradients in our implementations, whereas DeGMatch computes gradients with respect to all student parameters.
These values characterize our implementations and hardware rather than algorithm-independent complexity.
}
\label{tab:selection_compute}
\vspace{-10pt}
\end{table*}

\paragraph{Benchmark provenance.}
Our mathematical evaluation follows the core benchmark families used by Select2Reason~\citep{yang2026select2reason}, rather than reproducing its benchmark suite exactly. We evaluate on AIME24/25/26, AMC23, MATH500, Gaokao 2024/25, and HMMT Feb.\ 2025. For science and code, we follow the benchmark families used in OpenThoughts~\citep{guha2026openthoughts}, evaluating on GPQA-Diamond, the physics and chemistry portions of JEEBench, LiveCodeBench v6, and the Open-R1 CodeForces test split. Table~\ref{tab:evaluation_details} lists the corresponding benchmark citations and evaluation sizes. OpenThoughts additionally reports CodeElo~\citep{quan2025codeelo}; we omit it because its official evaluation protocol submits generated solutions directly to the Codeforces platform. We instead evaluate CodeForces solutions locally against the available official test cases.

\paragraph{Sampling variability.}
Repeated sampling is particularly important for small competition benchmarks. Across our evaluated runs, the standard error of the sampled accuracy estimate is approximately 0.8--1.2 percentage points for AIME under Avg@16 and 1.0--1.1 points for AMC under Avg@16. The corresponding ranges are approximately 0.6--0.8 points for GPQA-Diamond under Avg@8 and 0.5--1.1 points for JEEBench under Avg@4. As an additional diagnostic, splitting the 16 AIME generations into two disjoint groups of eight can produce accuracy estimates differing by as much as 5.8 points. We therefore use Avg@16 for the small mathematical competition benchmarks to reduce generation-sampling noise. These statistics characterize evaluation-time sampling variability and should not be interpreted as variation across independently trained models.

For AIME, AMC, HMMT, GPQA, MATH500, Gaokao, and LiveCodeBench, generation uses temperature $0.6$ and top-$p=0.95$. The current JEEBench and CodeForces harnesses use temperature $0.7$ and top-$p=1.0$.

% TODO before submission:
% Either standardize JEEBench/CodeForces decoding and re-evaluate,
% or retain the benchmark-specific settings and state this explicitly.

AIME and AMC answers are extracted from the final \texttt{\textbackslash boxed\{\}} expression. HMMT uses the MathArena answer extraction and verification procedure. MATH500 and Gaokao are graded using \texttt{math-verify}. GPQA-Diamond is scored using exact multiple-choice accuracy after deterministic option shuffling. LiveCodeBench and CodeForces generations are executed against benchmark test cases.

\subsection{Computational Resources and Cost}
\label{app:compute}

\paragraph{Hardware and DeGMatch selection cost.}
All experiments are conducted on NVIDIA RTX PRO 6000 Blackwell Max-Q GPUs with 96\,GB memory. Candidate-gradient features can be computed once for each student model and reused across target capabilities. For the primary student, computing projected gradients for the complete candidate pools requires approximately 58.3 GPU-hours for mathematics (93,733 examples), 16.4 GPU-hours for science (53,195 examples), and 35.1 GPU-hours for code (26,061 examples). Once the candidate features are available, computing gradients for a 50-example target-reference set requires at most 0.11 GPU-hours per target. Relevance-gate construction, PCA-based deconfounding, and final scoring require approximately five CPU minutes per mathematical target and are negligible relative to gradient extraction.

\paragraph{Selection-cost comparison.}
Table~\ref{tab:selection_compute} compares the approximate cost of constructing reusable selection features for the principal target-aware baselines and DeGMatch. All measurements correspond to our long-context implementations on RTX PRO 6000 GPUs. For LESS, GIST, and TAROT, the reported cost includes the auxiliary warm-up or adaptation used in our implementations together with candidate-gradient feature extraction. Candidate representations and warm-up checkpoints are reusable across targets for all applicable methods, so the comparison focuses on the one-time feature-construction cost rather than repeatedly charging shared computation for every target.

Under our implementation, DeGMatch requires roughly 29\% of the selection compute used by LESS or TAROT and is comparable in cost to GIST, despite using full-parameter rather than LoRA gradients. Its main practical distinction is that it requires no auxiliary warm-up or adaptation checkpoint: the reusable candidate representation is computed directly from the initial student. RDS+ remains substantially cheaper because it requires only a forward pass. For DeGMatch, adding a new target after candidate features have been computed requires only the small target-gradient computation and several minutes of CPU-side retrieval, deconfounding, and scoring. The same reuse principle applies to reusable candidate features and warm-up checkpoints of the reproduced baselines.

\begin{table}[t]
\centering
\footnotesize
\setlength{\tabcolsep}{3pt}
\renewcommand{\arraystretch}{1.06}

\begin{tabularx}{\columnwidth}{
@{}
>{\raggedright\arraybackslash}p{0.32\columnwidth}
*{3}{>{\centering\arraybackslash}X}
@{}
}
\toprule
\textbf{Target}
& \textbf{Input}
& \textbf{Output}
& \textbf{Cost} \\
\midrule

AIME
& 5.5K & 918.2K & \$1.98 \\

AMC
& 4.9K & 523.5K & \$1.31 \\

HMMT
& 5.6K & 1.08M & \$2.71 \\

MATH500
& 3.3K & 256.5K & \$0.64 \\

Gaokao
& 4.4K & 310.9K & \$0.78 \\

GPQA
& 9.4K & 486.0K & \$1.22 \\

JEEBench
& 6.1K & 280.7K & \$0.71 \\

LiveCodeBench
& 29.4K & 463.6K & \$1.18 \\

CodeForces
& 34.0K & 876.6K & \$2.22 \\

\midrule
\textbf{Total}
& \textbf{102.6K}
& \textbf{5.2M}
& \textbf{\$12.75} \\

\bottomrule
\end{tabularx}

\caption{
Token usage and monetary cost for generating verified target-reference trajectories.
Each target contains 50 reference examples, for 450 references in total.
Costs reflect the API pricing used at the time of generation.
}
\label{tab:reference_generation_cost}
\vspace{-10pt}
\end{table}

\paragraph{Training cost.}
Training cost depends substantially on the number of tokens contained in the selected subset. A 10K-example mathematical subset trained for five epochs typically requires approximately eight hours on the same eight-GPU node, corresponding to roughly 60--70 GPU-hours per run. Representative DeGMatch runs require approximately 7.8--8.4 hours, compared with 6.6 hours for Random and 9.5 hours for Length. This variation reflects differences in the total number and length of selected training tokens rather than differences in the optimization procedure. Science and code experiments use smaller example budgets and correspondingly shorter training runs.

\paragraph{Target-reference generation cost.}
Constructing the target-reference sets requires generating verified reasoning trajectories for the sampled reference problems. Table~\ref{tab:reference_generation_cost} reports the token usage and monetary cost incurred during this process. Across all nine target capabilities, generating 450 verified reference trajectories uses approximately 5.20M output tokens and costs \$12.75 in total, or about \$1.42 per target on average. Thus, the monetary cost of constructing the small target-reference sets is modest relative to subsequent selection and fine-tuning compute. Reported monetary costs correspond to the API prices in effect when the reference trajectories were generated.

\section{Diagnosing Target-Conditioned Gradient Matching}
\label{app:deconfounding_analysis}

We provide additional analysis of the gradient-matching limitation motivating the deconfounding step in Section~\ref{sec:deconfounded_gradient}. We examine two questions: (1) whether reasoning gradients contain dominant structure shared across the candidate population, and (2) whether this structure makes raw target-conditioned matching insufficiently sensitive to the supplied target references. We then analyze how candidate-population deconfounding changes this behavior.

\paragraph{Connection to common-component removal.}
\citet{mu2017allbuttop} observe that learned representations can contain a nonzero population mean and a small number of dominant principal directions shared broadly across examples. Their All-but-the-Top transformation centers the representation population and removes its leading principal subspace:
\begin{equation}
    x_i'
    =
    (I-U_DU_D^\top)(x_i-\mu_x).
    \label{eq:abtt_transform}
\end{equation}
The key geometric observation is that directions explaining substantial population-level variation need not be the directions that best distinguish individual examples under similarity-based comparison. DeGMatch considers the analogous issue for student gradients: whether dominant structure shared across candidate trajectories can overwhelm the variation that distinguishes one target capability from another.

\begin{table}[t]
\centering
\scriptsize
\setlength{\tabcolsep}{3.5pt}
\renewcommand{\arraystretch}{1.08}

\begin{tabular}{lrrrrr}
\toprule
\textbf{Target}
& \textbf{Top-1}
& \textbf{Top-5}
& \textbf{Top-10}
& \textbf{Top-50}
& \textbf{Top-200} \\
\midrule
AIME    & 0.343 & 0.732 & 0.794 & 0.873 & 0.917 \\
AMC     & 0.355 & 0.747 & 0.806 & 0.881 & 0.922 \\
HMMT    & 0.326 & 0.717 & 0.782 & 0.867 & 0.913 \\
MATH500 & 0.391 & 0.773 & 0.826 & 0.894 & 0.930 \\
Gaokao  & 0.385 & 0.769 & 0.822 & 0.892 & 0.929 \\
\bottomrule
\end{tabular}

\caption{
Cumulative variance explained by the leading principal
directions of centered candidate gradients in each
target-specific relevance gate.
}
\label{tab:pc_variance}
\vspace{-10pt}
\end{table}

\begin{figure*}[t]
    \centering
    \includegraphics[width=0.96\textwidth]{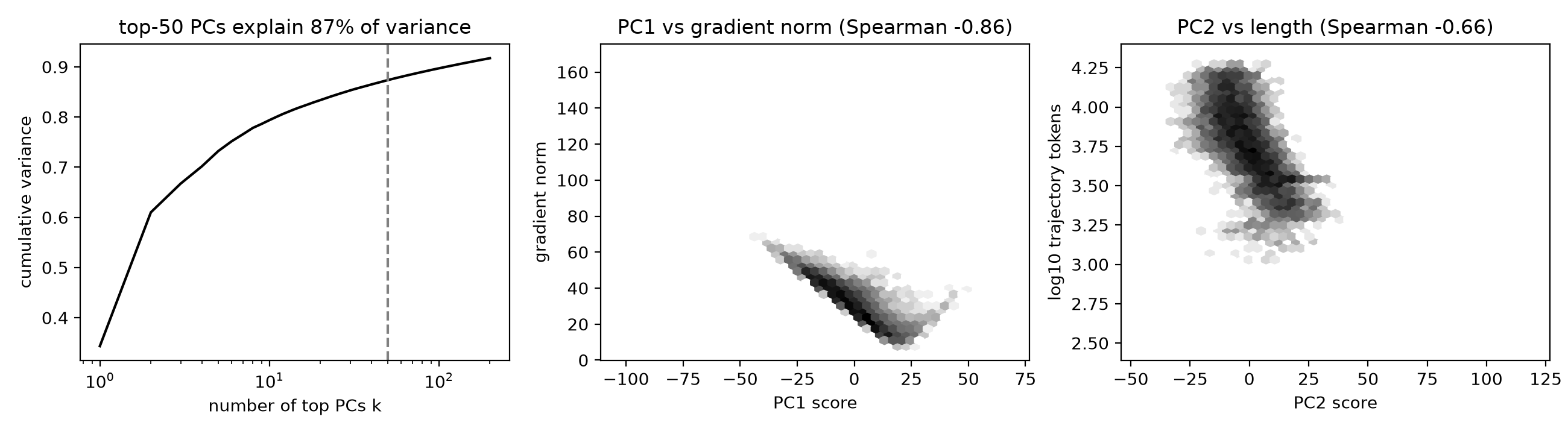}

    \vspace{3pt}

    \includegraphics[width=0.70\textwidth]{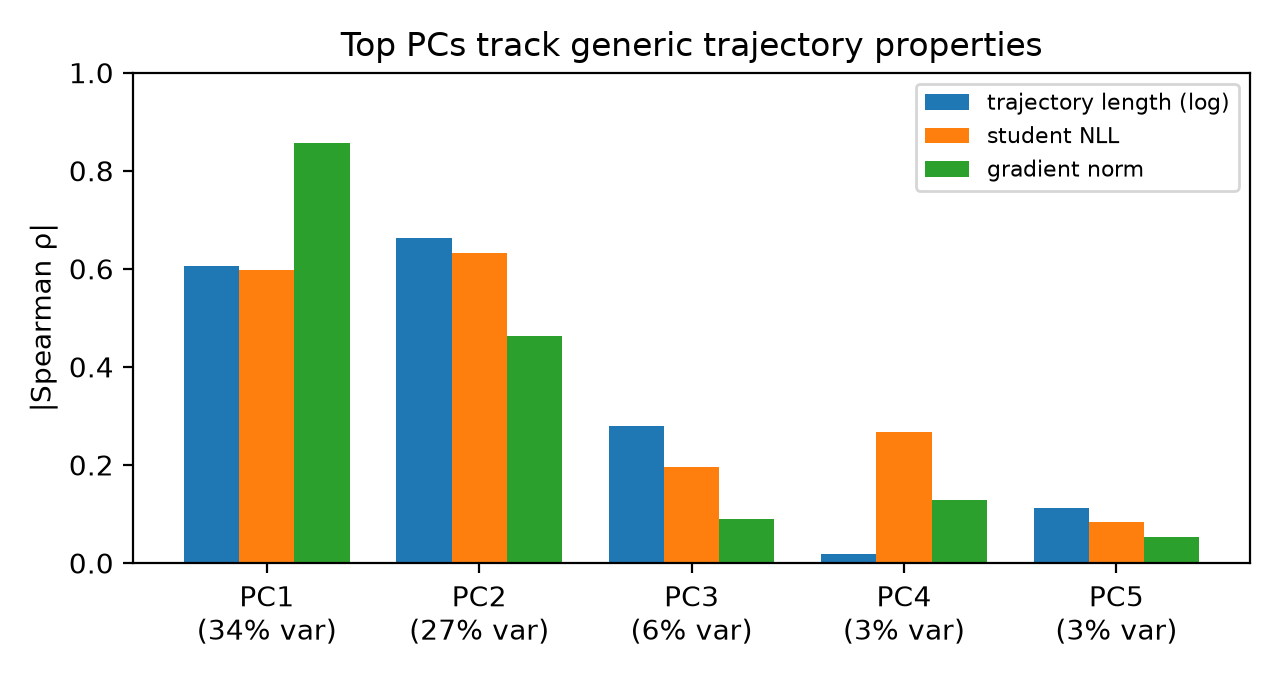}

    \caption{
    \textbf{Candidate gradients exhibit strongly concentrated population-level structure.}
    Top: cumulative variance explained by the leading principal components of centered gradient sketches in the 30K AIME relevance gate, together with representative associations between PC1 and gradient norm and between PC2 and trajectory length.
    Bottom: absolute Spearman correlations between the first five PC scores and trajectory length, student NLL, and gradient norm.
    The top 50 components explain 87.3\% of the variance; PC1 is strongly associated with gradient norm ($|\rho|=0.86$), and PC2 with trajectory length ($|\rho|=0.66$).
    }
    \label{fig:gradient_population_structure}
    \vspace{-10pt}
\end{figure*}

\paragraph{Why dominant components can reduce target sensitivity.}
Population centering removes the common mean offset, whereas principal-component removal suppresses dominant variation around that mean. These operations address distinct aspects of the gradient geometry, and we evaluate their effects separately.
Let $U\in\mathbb{R}^{d\times k}$ denote a dominant subspace of the centered candidate gradients. We can decompose a centered candidate gradient and target gradient as
\begin{equation}
    \bar g_i = Ua_i+r_i,
    \qquad
    \bar g_j^t = Ub_j^t+s_j^t,
\end{equation}
where
\begin{equation}
    U^\top r_i=0,
    \qquad
    U^\top s_j^t=0.
\end{equation}
Their inner product therefore decomposes as
\begin{equation}
    \bar g_i^\top \bar g_j^t
    =
    a_i^\top b_j^t
    +
    r_i^\top s_j^t.
    \label{eq:gradient_inner_decomposition}
\end{equation}
Correspondingly,
\begin{equation}
    \cos(\bar g_i,\bar g_j^t)
    =
    \frac{
        a_i^\top b_j^t+r_i^\top s_j^t
    }{
        \sqrt{\|a_i\|_2^2+\|r_i\|_2^2}
        \sqrt{\|b_j^t\|_2^2+\|s_j^t\|_2^2}
    }.
    \label{eq:gradient_cosine_decomposition}
\end{equation}

If a high-energy subspace captures broad properties shared across candidate
and reference trajectories, its contribution $a_i^\top b_j^t$ can dominate
pairwise similarity even when those properties are not specific to the target
capability of interest. In this case, references that match such broad
trajectory properties can induce candidate rankings similar to those produced
by the true target references, thereby obscuring finer target-conditioned
variation in the matching geometry.

After removing the dominant candidate subspace,
\begin{equation}
    P_\perp\bar g_i=r_i,
    \qquad
    P_\perp\bar g_j^t=s_j^t,
\end{equation}
and cosine matching depends only on the residual geometry:
\begin{equation}
    \cos(P_\perp\bar g_i,P_\perp\bar g_j^t)
    =
    \frac{r_i^\top s_j^t}
         {\|r_i\|_2\|s_j^t\|_2}.
    \label{eq:residual_cosine}
\end{equation}

\begin{table*}[t]
\centering
\small
\setlength{\tabcolsep}{4pt}
\renewcommand{\arraystretch}{1.06}

\begin{tabularx}{\textwidth}{
@{}
>{\raggedright\arraybackslash}p{0.23\textwidth}
*{5}{>{\centering\arraybackslash}X}
>{\centering\arraybackslash}p{0.18\textwidth}
@{}
}
\toprule
\textbf{Property}
& \textbf{PC1}
& \textbf{PC2}
& \textbf{PC3}
& \textbf{PC4}
& \textbf{PC5}
& \textbf{\# PCs with $|\rho|>0.3$} \\
\midrule

Log trajectory length
& 0.61 & 0.66 & 0.28 & 0.02 & 0.11 & 2 \\

Student NLL
& 0.60 & 0.63 & 0.20 & 0.27 & 0.08 & 2 \\

Gradient norm
& 0.86 & 0.46 & 0.09 & 0.13 & 0.05 & 2 \\

Olympiad source
& 0.01 & 0.12 & 0.07 & 0.12 & 0.06 & 0 \\

Algebra
& 0.16 & 0.22 & 0.18 & 0.19 & 0.02 & 0 \\

Geometry
& 0.08 & 0.10 & 0.10 & 0.15 & 0.35 & 2 \\

\bottomrule
\end{tabularx}

\caption{
Absolute Spearman correlations between scores along the leading gradient PCs and trajectory properties.
The final column counts components among the first 50 for which $|\rho|>0.3$.
}
\label{tab:pc_correlations}
\vspace{-10pt}
\end{table*}

This argument does not require the removed directions to be uninformative for optimization. A direction can contain substantial gradient energy while still being poorly suited for distinguishing target identity if it is shared broadly across candidates and targets. DeGMatch therefore removes these directions only from the geometry used for selection; training on the selected trajectories uses the original supervision and the full model gradients.

This argument yields two empirical predictions. First, candidate gradients
should contain a high-energy population-level subspace that strongly determines
raw gradient rankings. Second, if this subspace captures broad trajectory
properties rather than target content alone, non-target references matched on
such properties should reproduce much of the raw candidate ordering, and this
dependence should be reduced specifically by removing the dominant directions.
We test these predictions below.

\subsection{Dominant Shared Structure in Candidate Gradients}
\label{app:dominant_directions}

We first test whether the concentration of reasoning gradients into a small number of dominant directions is specific to a particular target-conditioned candidate population. Table~\ref{tab:pc_variance} reports the cumulative variance explained by the leading principal components after centering each of the five mathematical relevance gates by its own candidate-population mean. The spectra are consistently concentrated across targets: the first 10 components explain 78.2--82.6\% of the variance, and the first 50 explain 86.7--89.4\%. Thus, the dominant-spectrum phenomenon observed for AIME is not specific to a single relevance gate.

To characterize what varies along these directions, Table~\ref{tab:pc_correlations} reports absolute Spearman correlations between scores along the leading PCs and several trajectory-level properties. PC1 is strongly associated with gradient norm ($|\rho|=0.86$), while PC1 and PC2 are also associated with trajectory length and student NLL. In contrast, source and problem-type indicators exhibit substantially weaker associations with the dominant components. These correlations are descriptive and do not imply that trajectory length, NLL, or gradient norm causes the dominant structure.

Figure~\ref{fig:gradient_population_structure} provides a complementary visualization of these diagnostics. The spectrum is strongly concentrated, while the scatter plots and correlation summary show that the leading components are associated with broad trajectory-level quantities such as gradient magnitude, sequence length, and student NLL. Together with Tables~\ref{tab:pc_variance} and~\ref{tab:pc_correlations}, this indicates that raw gradient similarity is computed in a space containing substantial candidate-population structure shared across many trajectories.

\begin{table}[t]
\centering
\footnotesize
\setlength{\tabcolsep}{1.5pt}
\renewcommand{\arraystretch}{1.06}

\begin{tabularx}{\columnwidth}{
@{}
>{\raggedright\arraybackslash}p{0.36\columnwidth}
*{4}{>{\centering\arraybackslash}X}
@{}
}
\toprule
\textbf{Set}
& \textbf{$k=10$}
& \textbf{$k=25$}
& \textbf{$k=50$}
& \textbf{\mbox{$k=100$}} \\
\midrule

Random examples
& 0.774 & 0.827 & 0.861 & 0.884 \\

AIME references
& 0.658 & 0.720 & 0.777 & 0.817 \\

AMC references
& 0.687 & 0.749 & 0.805 & 0.841 \\

HMMT references
& 0.654 & 0.714 & 0.781 & 0.819 \\

MATH500 references
& 0.809 & 0.856 & 0.886 & 0.906 \\

Gaokao references
& 0.795 & 0.835 & 0.865 & 0.889 \\

\bottomrule
\end{tabularx}

\caption{
Fraction of centered-gradient energy contained in the leading candidate-gradient subspace.
The gate statistics are computed from 50 randomly sampled gate examples.
}
\label{tab:target_pc_energy}
\vspace{-10pt}
\end{table}

We additionally compare how much gradient energy candidate and target examples allocate to the dominant candidate subspace. Table~\ref{tab:target_pc_energy} reports the fraction of squared centered-gradient norm captured by its first $k$ principal directions. At $k=50$, 86.1\% of the energy of randomly sampled gate gradients lies in this subspace, compared with 77.7\%, 80.5\%, and 78.1\% for AIME, AMC, and HMMT target references, respectively.

\begin{table}[t]
\centering
\small
\setlength{\tabcolsep}{4pt}
\renewcommand{\arraystretch}{1.08}

\begin{tabular}{lrrr}
\toprule
\textbf{PC range}
& \textbf{Across targets}
& \textbf{Pool-split ref.}
& \textbf{Random} \\
\midrule
1--5    & 0.93 & 0.94 & 0.0003 \\
1--50   & 0.79 & 0.87 & 0.0031 \\
51--100 & 0.45 & 0.60 & 0.0031 \\
\bottomrule
\end{tabular}

\caption{
Normalized subspace overlap across target-conditioned
candidate populations after excluding candidates shared
between relevance gates.
The pool-split reference uses non-overlapping random subsets
of the mathematical pool and reflects finite-sample stability.
}
\label{tab:shared_candidate_subspace}
\vspace{-10pt}
\end{table}

\begin{figure*}[t]
    \centering
    \includegraphics[width=\textwidth]{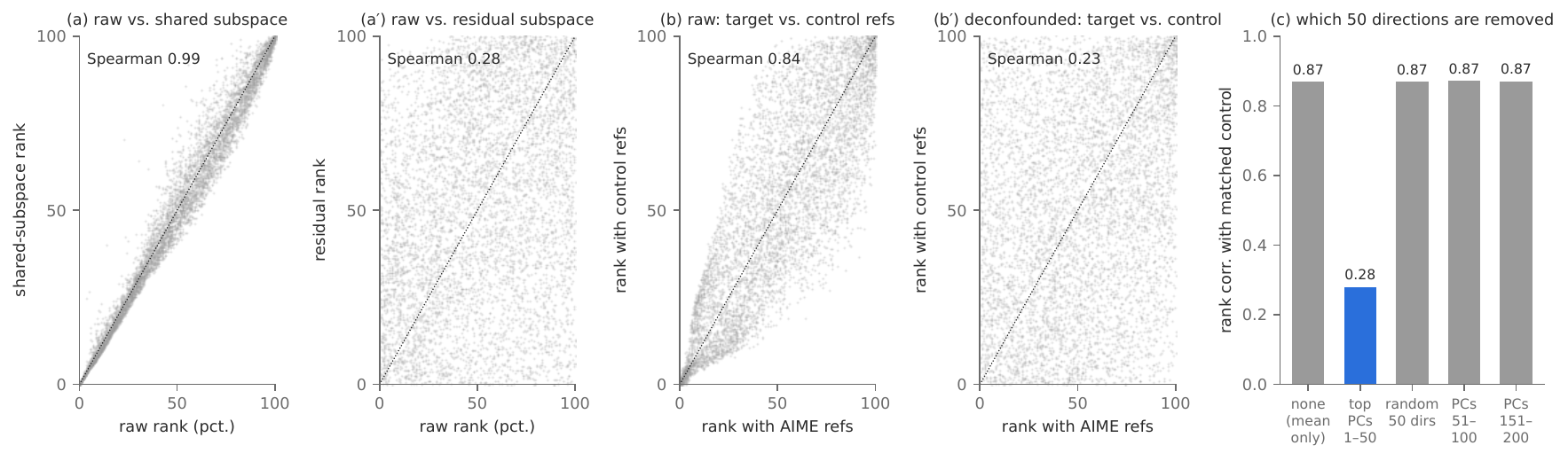}
    \caption{
    \textbf{Why DeGMatch removes the dominant candidate-gradient directions.}
    (a) The raw AIME ranking is almost completely reproduced using only the
    dominant population component (the mean direction and top 50 centered
    candidate PCs), whereas
    (a$'$) the orthogonal residual alone produces a substantially different
    ranking.
    (b--b$'$) A length- and NLL-matched non-target reference set induces a
    ranking highly correlated with the AIME ranking in the raw gradient space,
    but this correspondence is sharply reduced after dominant-component
    removal.
    (c) The effect is specific to the leading principal directions:
    after population centering, removing the top 50 PCs reduces the
    matched-control rank correlation from $0.87$ to $0.28$, whereas removing
    50 random directions or lower-ranked PC blocks leaves it essentially
    unchanged.
    }
    \label{fig:why_remove_pcs}
    \vspace{-10pt}
\end{figure*}

We next ask whether the dominant directions recur across
different target-conditioned candidate populations.
A direct comparison of the relevance gates is potentially
confounded by shared candidates, so we construct a stricter
control in which each target retains only candidates that
appear in no other target's gate.
We then recompute PCA independently on these mutually
exclusive candidate sets.

For two $k$-dimensional principal subspaces with orthonormal
bases $U$ and $V$, we measure normalized subspace overlap as
\[
    \operatorname{overlap}(U,V)
    =
    \frac{1}{k}\lVert U^\top V\rVert_F^2.
\]
This quantity equals one for identical subspaces and approaches
zero for unrelated low-dimensional subspaces.

As shown in Table~\ref{tab:shared_candidate_subspace}, the dominant directions remain strongly aligned even under this zero-overlap construction. The first five components have a mean cross-target subspace overlap of 0.93, nearly matching the 0.94 overlap obtained from non-overlapping random subsets of the same underlying mathematical pool. The first 50 components also remain substantially aligned (0.79 versus a 0.87 pool-split reference), whereas components 51--100 are markedly less shared (0.45 versus 0.60). Thus, the strongest population-level directions recur across target-conditioned candidate populations even when their underlying examples do not overlap. We emphasize that these candidate gates are not themselves ground-truth samples from the downstream tasks; this analysis characterizes shared structure in the candidate populations induced by different target references.

Together, these results establish two properties of the candidate-gradient geometry: its variance is strongly concentrated in a small number of leading directions across targets, and these dominant directions recur across distinct target-conditioned candidate populations. We next examine whether this shared high-variance structure interferes with target-conditioned gradient matching.

\subsection{Why Remove the Dominant Principal Directions?}
\label{app:why_remove_pcs}

The preceding analysis establishes that candidate gradients contain strong
population-level structure, but high variance alone does not justify removing
the corresponding directions. We therefore ask a more direct question:
does this structure cause target-conditioned gradient matching to respond to
broad trajectory properties rather than to the target references themselves?

\paragraph{A matched non-target control.}
We construct a negative control for the AIME target by replacing each AIME
reference with a non-target trajectory outside the AIME relevance gate that is
closely matched in trajectory length and student NLL. We repeat this
construction with three random seeds. The candidate population is held fixed
to the same 30K AIME relevance gate used by DeGMatch. Unless otherwise noted,
the diagnostics in this subsection use the directional matching score $d_i$,
so that they isolate the similarity geometry affected by deconfounding.

\begin{table*}[t]
\centering
\small
\setlength{\tabcolsep}{4pt}
\renewcommand{\arraystretch}{1.08}

\begin{tabularx}{\textwidth}{
@{}
>{\raggedright\arraybackslash}p{0.32\textwidth}
*{5}{>{\centering\arraybackslash}X}
@{}
}
\toprule
\textbf{Representation / correction}
& \textbf{Matched-control $\rho$}
& \textbf{Random-ref. $\rho$}
& \textbf{Overlap w/ deconf.}
& \textbf{Top-3K tokens}
& \textbf{$\rho$(score, len.)} \\
\midrule

Raw direction
& 0.86
& 0.58
& 0.18
& 28.5M
& +0.72 \\

Score residualization (linear)
& 0.79
& 0.62
& 0.02
& 13.2M
& $-0.17$ \\

Score residualization (cubic)
& 0.70
& 0.56
& 0.02
& 10.4M
& $-0.11$ \\

Gradient residualization
& 0.78
& 0.73
& 0.12
& 17.2M
& $-0.16$ \\

\textbf{Mean + PC50}
& \textbf{0.28}
& \textbf{0.08}
& --
& 30.8M
& +0.34 \\

Covariate residualization + PC50
& \textbf{0.24}
& 0.09
& 0.63
& 30.0M
& +0.32 \\

\bottomrule
\end{tabularx}

\caption{
Comparison with explicit residualization of measured trajectory covariates
within the fixed 30K AIME relevance gate.
\emph{Matched-control $\rho$} is the Spearman correlation between rankings
induced by the AIME references and by non-target references matched in
trajectory length and student NLL.
\emph{Random-ref.\ $\rho$} reports correlation with rankings induced by random
reference sets.
\emph{Overlap w/ deconf.} is the fraction of the top-3K subset shared with the
default mean-centered, top-50-PC-removed directional ranking.
The regression variants use log trajectory length, student NLL, and log
gradient norm as covariates.
}
\label{tab:covariate_residualization}
\vspace{-10pt}
\end{table*}

Raw gradient matching fails this negative-control test: the ranking induced by
the true AIME references is largely reproduced by the matched non-target
references (mean Spearman $\rho=0.86$). Thus, much of the raw candidate
ordering can be recovered from references that match broad trajectory
statistics without specifying the intended target capability.

\paragraph{Raw rankings are dominated by population structure.}
To identify the source of this behavior, we decompose each gradient into the
component contained in the dominant candidate-population space and its
orthogonal residual. The dominant component consists of the population mean
together with the top 50 principal directions estimated from the centered
30K gate gradients. Across the five mathematical targets, rankings computed
using only this dominant component correlate with the corresponding raw
rankings at $\rho=0.96$--$0.99$; for AIME, the correlation is $0.99$ and the
top-10\% overlap is $0.85$. In contrast, rankings computed from the residual
component alone correlate with the raw rankings at only $0.01$--$0.28$.
The same dominance is observed with randomly sampled reference trajectories,
for which the dominant component still reproduces the raw ranking with
$\rho=0.96$.

Figure~\ref{fig:why_remove_pcs} visualizes this decomposition and the
negative-control test. The result does not imply that the leading components
contain no target-related information. Rather, it shows that the raw matching
geometry is dominated by population-level structure through which broad
reference properties can strongly determine the candidate ordering.

\paragraph{The effect is specific to the leading directions.}
A possible alternative explanation is that matching improves simply because
50 dimensions are removed. We test this by removing the same number of
directions from different parts of the gradient space after population
centering. Removing 50 random orthogonal directions, PCs 51--100, or
PCs 151--200 leaves the matched-control correlation essentially unchanged at
$\rho\approx0.87$. Only removal of PCs 1--50 reduces it substantially, to
$0.28$. The same intervention also reduces correlation with rankings induced
by random reference sets. These controls show that the effect is not a generic
consequence of dimensionality reduction, but is specific to suppressing the
dominant candidate-population directions.

\begin{figure*}[t]
    \centering
    \includegraphics[width=0.82\textwidth]
    {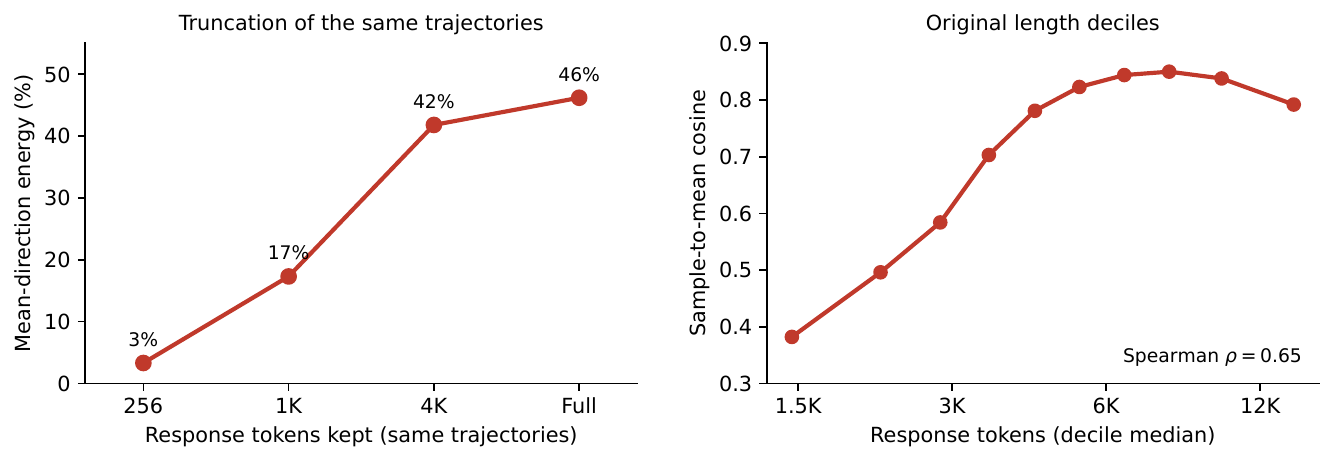}
    \caption{
    \textbf{Shared gradient structure becomes more prominent with
    reasoning-trajectory length.}
    Left: mean-direction energy after recomputing gradients for the same
    reasoning subset truncated to 256, 1,024, and 4,096 response tokens,
    together with the full trajectories.
    Right: sample-to-mean gradient cosine across deciles of the original
    reasoning-trajectory length distribution (20,000 trajectories).
    The truncation analysis holds the underlying examples fixed while varying
    the observed response length, whereas the decile analysis measures the
    corresponding association in the original data.
    }
    \label{fig:reasoning_length_structure}
    \vspace{-10pt}
\end{figure*}

\begin{figure*}[t]
    \centering
    \includegraphics[width=0.66\textwidth]{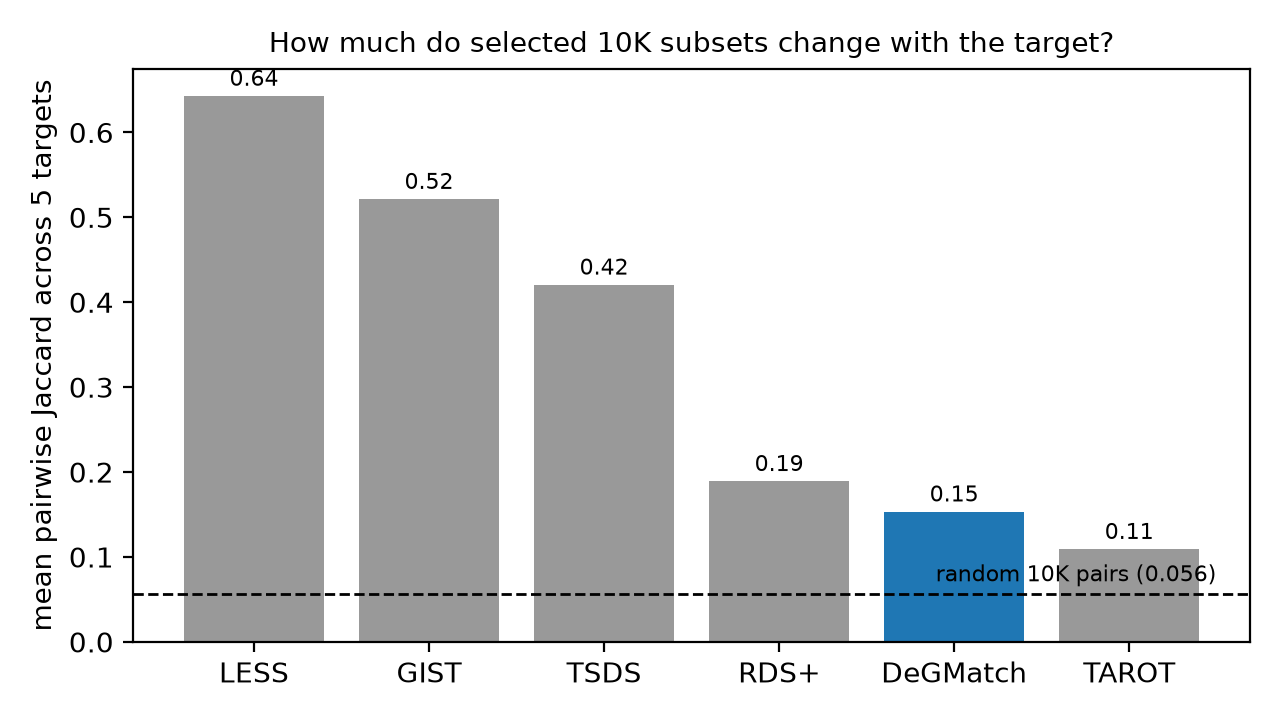}

    \vspace{5pt}

    \includegraphics[width=0.98\textwidth]{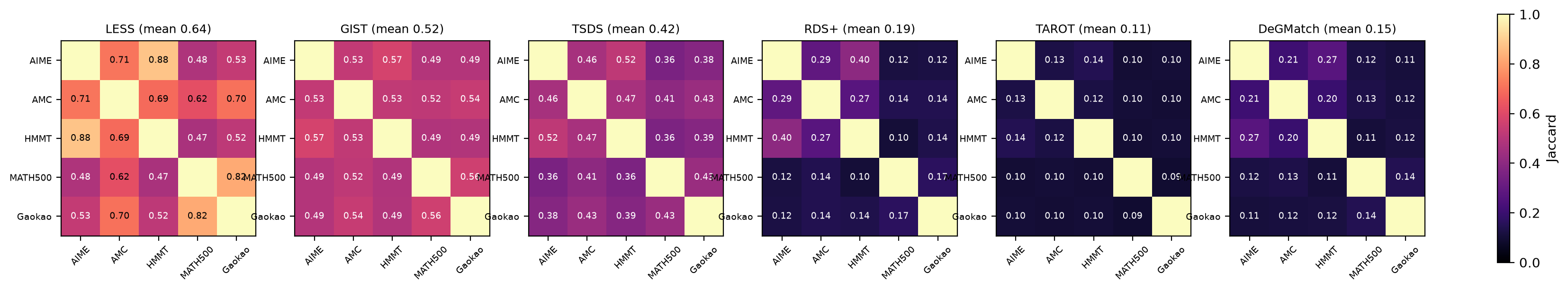}

    \caption{
    \textbf{Target sensitivity varies substantially across target-aware selectors.}
    Top: mean pairwise Jaccard similarity among the 10K subsets independently selected for AIME, AMC, HMMT, MATH500, and Gaokao.
    The dashed line denotes the expected Jaccard similarity between two independently sampled 10K subsets from the 93,733-example mathematical pool.
    Bottom: pairwise Jaccard matrices across the five targets.
    Higher overlap indicates that the selected subset changes less when the supplied target references change.
    }
    \label{fig:baseline_target_sensitivity}
    % \vspace{-10pt}
\end{figure*}

\subsection{Can Measured Covariates Explain the Confounding?}
\label{app:covariate_residualization}

The analyses above show that the dominant gradient directions are associated
with broad trajectory properties such as sequence length, student NLL, and
gradient magnitude. This raises a simpler alternative: rather than estimating
and removing dominant population structure, could we directly regress out
these measured properties?

We test this possibility in the same fixed 30K AIME relevance gate.
We consider three alternatives to principal-component removal.
First, we residualize the raw directional matching score using a linear
regression on log trajectory length, student NLL, and log gradient norm.
Second, we use a cubic polynomial expansion with interaction terms.
Third, we regress the same covariates directly from the gradient
representations before computing directional matching.
Table~\ref{tab:covariate_residualization} reports the resulting diagnostics.

Explicit covariate correction substantially reduces the direct association
between the matching score and trajectory length: the score--length
correlation changes from $0.72$ for raw matching to between $-0.11$ and
$-0.17$ after score- or gradient-level residualization.
Nevertheless, the negative-control dependence remains strong.
The length- and NLL-matched non-target references still reproduce the
resulting candidate rankings with correlations of $0.70$--$0.79$, compared
with $0.28$ after removing the dominant candidate PCs.

This difference indicates that eliminating the marginal dependence on a few
measured trajectory statistics is not sufficient to remove the population
structure that dominates raw matching.
Consistent with this interpretation, removing the top 50 PCs after
gradient-level covariate residualization further lowers the matched-control
correlation to $0.24$, only slightly below PC removal alone.
Moreover, the top-50 principal subspaces estimated before and after covariate
residualization have a projection overlap of $0.97$.
Thus, the dominant candidate subspace appears to subsume variation associated
with length, NLL, and gradient scale while also capturing additional shared
multidimensional structure that is not specified by these measured
covariates.

\begin{table}[t]
\centering
\footnotesize
\setlength{\tabcolsep}{3.5pt}
\renewcommand{\arraystretch}{1.08}

\begin{tabularx}{\columnwidth}{
@{}
>{\raggedright\arraybackslash}X
ccc
@{}
}
\toprule
& \textbf{IT}
& \textbf{Short Math}
& \textbf{R1 Reasoning} \\
\midrule

Median response tokens
& 33 & 159 & 4,826 \\

Mean-direction energy
& 0.016 & 0.100 & \textbf{0.438} \\

Sample--mean cosine
& $-0.02$ & 0.40 & \textbf{0.71} \\

Raw random-ref.\ $\rho$
& 0.35 & -- & \textbf{0.75} \\

\bottomrule
\end{tabularx}

\caption{
Shared gradient structure across instruction-tuning examples, short
mathematical solutions, and long R1 reasoning trajectories
(4,000 examples each).
All gradients are computed using the same student model and CountSketch
representation.
\emph{Mean-direction energy} is the fraction of gradient energy contained in
the population mean direction, and \emph{sample--mean cosine} measures the
alignment of individual gradients with that direction.
\emph{Raw random-ref.\ $\rho$} is the Spearman correlation between rankings
induced by the target references and randomly sampled reference trajectories.
}
\label{tab:reasoning_shared_structure}
\vspace{-10pt}
\end{table}

\subsection{Shared Structure and Reasoning-Trajectory Length}
\label{app:reasoning_length_structure}

We finally examine whether the shared population structure is associated with
the long trajectories characteristic of reasoning post-training.
We compare instruction-tuning examples from the LESS data pool, short
mathematical solutions from MATH, and long R1 reasoning trajectories using the
same student model, gradient extraction procedure, and CountSketch
representation.

Table~\ref{tab:reasoning_shared_structure} shows that the clearest difference
appears in the shared mean direction rather than in the centered PCA spectrum.
For instruction-tuning examples, the mean direction accounts for only $1.6\%$
of gradient energy and the average sample-to-mean cosine is $-0.02$.
For long R1 reasoning trajectories, the corresponding values increase to
$43.8\%$ and $0.71$, respectively. Short mathematical solutions lie between
these settings, with $10.0\%$ mean-direction energy and a sample-to-mean cosine
of $0.40$, suggesting that the strong shared structure observed for R1
trajectories is not attributable to the mathematical domain alone.
Consistently, raw reasoning-gradient rankings are substantially more
reproducible under random reference sets than instruction-tuning rankings
($\rho=0.75$ vs.\ $0.35$).

\begin{table*}[t]
\centering
\small
\setlength{\tabcolsep}{4pt}
\renewcommand{\arraystretch}{1.06}

\begin{tabularx}{\textwidth}{
@{}
>{\raggedright\arraybackslash}p{0.20\textwidth}
>{\centering\arraybackslash}p{0.13\textwidth}
*{4}{Y}
>{\centering\arraybackslash}p{0.12\textwidth}
@{}
}
\toprule
\textbf{Representation}
& \textbf{Matched-control $\rho$}
& \textbf{AMC}
& \textbf{HMMT}
& \textbf{MATH500}
& \textbf{Gaokao}
& \textbf{$\rho$(score, len.)} \\
\midrule

Raw gradients
& 0.86
& 0.92 & 0.96 & 0.90 & 0.89
& 0.78 \\

Mean-centered
& 0.87
& 0.68 & 0.88 & 0.45 & 0.45
& 0.58 \\

Mean + PC10 removal
& 0.39
& 0.75 & 0.78 & 0.62 & 0.66
& 0.56 \\

Mean + PC25 removal
& 0.36
& 0.74 & 0.79 & 0.58 & 0.59
& 0.63 \\

\textbf{Mean + PC50 removal}
& \textbf{0.28}
& \textbf{0.72} & \textbf{0.78} & \textbf{0.56} & \textbf{0.59}
& 0.69 \\

Mean + PC100 removal
& 0.24
& 0.71 & 0.75 & 0.57 & 0.58
& 0.69 \\

Mean + PC200 removal
& 0.20
& 0.69 & 0.74 & 0.55 & 0.57
& 0.69 \\

\bottomrule
\end{tabularx}

\caption{
Target-specificity diagnostics on the AIME relevance gate.
\emph{Matched-control $\rho$} is the Spearman correlation obtained after replacing the AIME references with a length- and NLL-matched non-target control.
The four middle columns report top-10K intersection fractions after replacing the AIME references with another mathematical target.
}
\label{tab:target_specificity_deconf}
\vspace{-10pt}
\end{table*}

To examine the association with trajectory length more directly, we recompute
gradients for a fixed subset of up to 2,000 reasoning examples after truncating
their responses to 256, 1,024, and 4,096 tokens.
As shown in Figure~\ref{fig:reasoning_length_structure}, mean-direction energy
increases from $3\%$ at 256 tokens to $17\%$ at 1,024 tokens and $42\%$ at
4,096 tokens, reaching $46\%$ for the full trajectories.
The same pattern appears in the unmodified reasoning data: across
trajectory-length deciles, sample-to-mean alignment increases with length and
plateaus for the longest trajectories (Spearman $\rho=0.65$).

Together, these results are consistent with shared gradient structure becoming
more prominent as reasoning trajectories grow longer. We treat this analysis
as evidence consistent with a length-related mechanism rather than as causal
identification of trajectory length as the sole source of the shared
population structure.

\subsection{Target Sensitivity of Existing Target-Aware Selectors}
\label{app:existing_target_sensitivity}

We first examine the selected subsets produced by existing target-aware methods. For each selector, we independently obtain 10K-example subsets for AIME, AMC, HMMT, MATH500, and Gaokao and compute the pairwise Jaccard similarity
\[
J(S_t,S_{t'})=
\frac{|S_t\cap S_{t'}|}{|S_t\cup S_{t'}|}
\]
between the resulting target-specific subsets. We additionally include TSDS~\citep{liu2024tsds} as an auxiliary target-aware comparison for this diagnostic.

\begin{figure*}[t]
    \centering
    \includegraphics[width=0.98\textwidth]{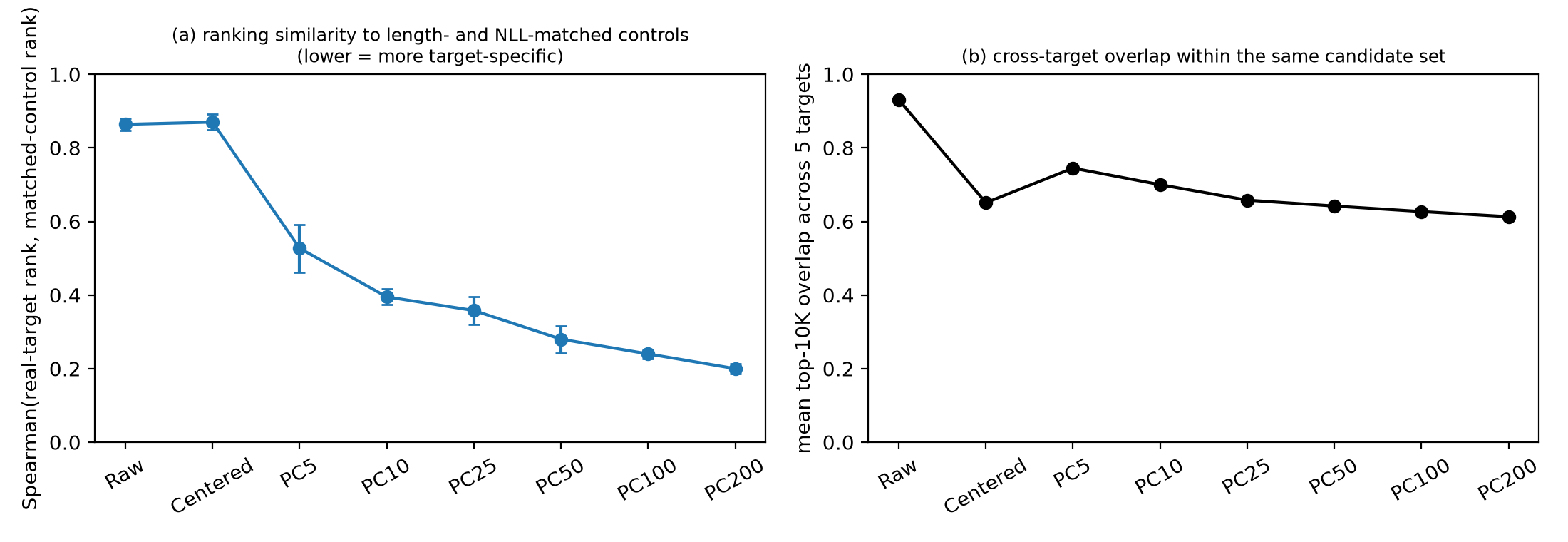}

    \vspace{5pt}

    \includegraphics[width=0.98\textwidth]{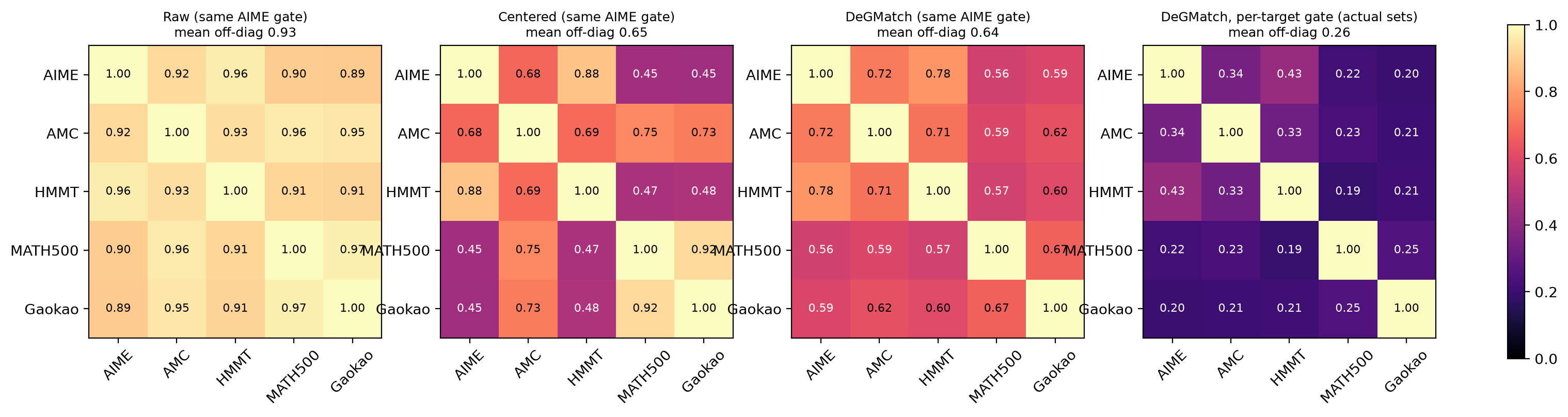}

    \caption{
    \textbf{Removing dominant candidate directions increases target sensitivity.}
    Top left: Spearman correlation between the ranking induced by the AIME references and rankings induced by length- and NLL-matched non-target controls as the gradient representation changes (mean $\pm$ s.d.\ over three draws).
    Mean-centering leaves the matched-control correlation essentially unchanged, whereas principal-component removal sharply reduces it.
    Top right: mean pairwise intersection fraction among the top-10K subsets induced by five mathematical targets when all targets score the same 30K AIME candidate set.
    Bottom: pairwise intersection fractions for raw gradients, mean-centered gradients, mean plus top-50-PC removal, and the final DeGMatch selections using target-specific relevance gates.
    }
    \label{fig:deconf_target_sensitivity}
    \vspace{-10pt}
\end{figure*}

Figure~\ref{fig:baseline_target_sensitivity} shows that target sensitivity varies substantially across methods. LESS, GIST, and TSDS produce highly overlapping subsets across targets, with mean pairwise Jaccard similarities of 0.64, 0.52, and 0.42, respectively. For comparison, two independently sampled uniform 10K subsets from the 93,733-example mathematical pool have an expected Jaccard similarity of approximately 0.056. The corresponding pairwise matrices show that the high averages are not driven by a single target pair: LESS and GIST in particular retain substantial overlap across most target combinations.

RDS+ and TAROT exhibit stronger differentiation across targets, with mean Jaccard similarities of 0.19 and 0.11, respectively, while DeGMatch yields 0.15. Thus, the diagnostic does not imply that all existing target-aware methods are insensitive to target identity. Rather, it shows that the extent to which selected data respond to the supplied target varies substantially across existing methods.

Importantly, lower overlap is not itself a measure of selection quality. TAROT produces even lower cross-target overlap than DeGMatch, yet does not outperform the strongest target-agnostic baselines in downstream evaluation. We therefore use cross-target overlap only as a diagnostic of how strongly a selector responds to target identity, rather than as an objective that should be minimized.

% \subsection{Deconfounding Increases Target Specificity}
\subsection{Additional Target-Sensitivity Diagnostics}
\label{app:deconf_target_specificity}

As a complementary diagnostic, we also examine how rankings and selected
subsets vary across the five mathematical target reference sets.
These cross-target comparisons should not be interpreted as a direct measure
of selection quality: lower overlap is not intrinsically better, and different
targets may legitimately share useful training examples.

\begin{figure*}[t]
    \centering
    \includegraphics[width=\textwidth]
    {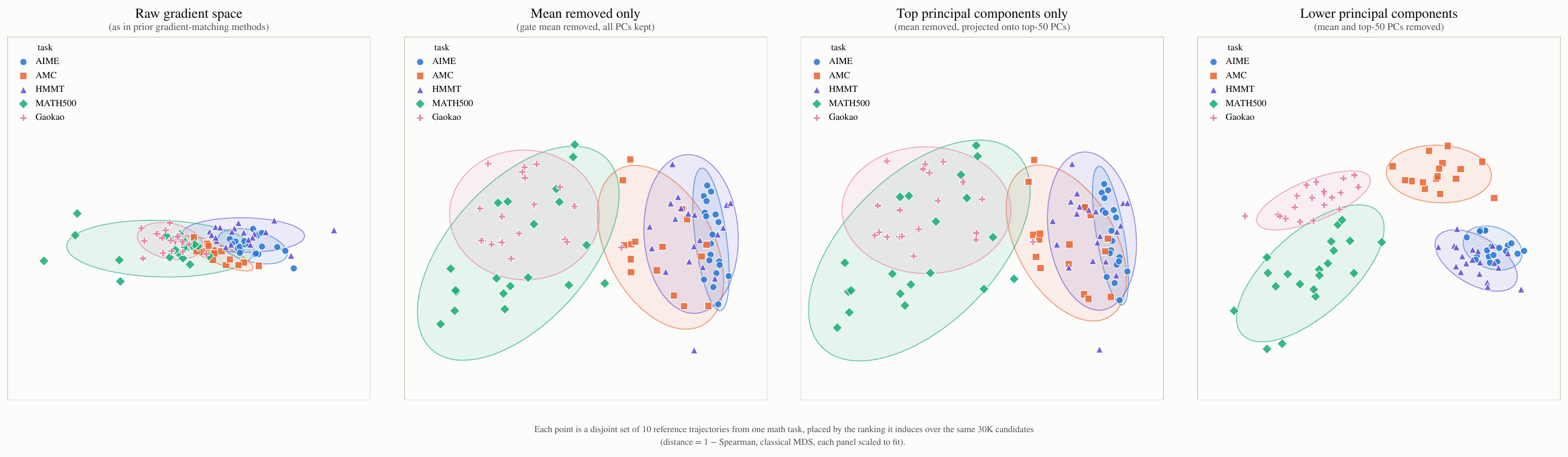}
    \caption{
    \textbf{Dominant-component removal reveals finer
    target-conditioned structure in candidate rankings.}
    Each point represents the ranking induced by a disjoint
    10-example reference subset over the same fixed 30K
    candidate population.
    From left to right: raw gradients, population centering
    alone, projection onto the top 50 principal components
    after centering, and the residual representation after
    removing the population mean and top 50 components.
    Distances are $1-\mathrm{Spearman}$ and are visualized
    using classical MDS; each panel is scaled independently
    for visualization.
    Quantitative target-identification results are computed
    in the original ranking space rather than from the
    two-dimensional projection.
    }
    \label{fig:reference_subset_structure}
    \vspace{-10pt}
\end{figure*}

Table~\ref{tab:target_specificity_deconf} reports two complementary diagnostics. Matched-control $\rho$ is the Spearman correlation between the AIME-conditioned ranking and a ranking obtained after replacing the AIME references with a non-target control matched for trajectory length and student NLL. The remaining columns measure the intersection fraction between equal-size top-10K subsets,
\[
O(S_t,S_{t'})
=
\frac{|S_t\cap S_{t'}|}{10{,}000},
\]
after replacing the AIME references with AMC, HMMT, MATH500, or Gaokao references. This differs from the Jaccard similarity used in Section~\ref{app:existing_target_sensitivity}: here the denominator is the fixed selection budget rather than the union size. These quantities are diagnostics of target dependence rather than performance metrics; lower overlap is not intrinsically better.

Raw gradient matching is strikingly insensitive to target replacement. Replacing the AIME references with the matched non-target control preserves a ranking correlation of $0.86$, while replacing them with another mathematical target retains 89--96\% of the raw top-10K set. Population centering changes the cross-target rankings but leaves matched-control correlation essentially unchanged at $0.87$, indicating that mean removal alone does not resolve the target-insensitivity captured by this diagnostic.

Removing the dominant candidate principal directions produces a qualitatively different geometry. With $k=50$, matched-control correlation falls from 0.86 to 0.28, while the cross-target top-10K intersection fraction falls to 0.56--0.78. The resulting rankings also preserve non-uniform relationships among targets: AIME remains more similar to AMC and HMMT than to MATH500 or Gaokao. Thus, dominant-component removal makes the candidate ranking substantially more dependent on which target references are supplied, rather than merely applying a target-independent perturbation.

Figure~\ref{fig:deconf_target_sensitivity} visualizes this effect while holding the candidate population fixed. The left panel tracks ranking similarity to the matched non-target control as progressively more of the dominant candidate subspace is removed. Mean-centering alone leaves the correlation essentially unchanged, whereas removing the leading principal directions produces a sharp reduction. This distinction is important: although centering changes the overall similarity geometry, the dominant variation around the mean remains sufficient for a matched non-target reference set to induce nearly the same candidate ordering as the true target.

The right panel measures the mean pairwise top-10K overlap among all five mathematical targets while every target scores the same 30K AIME gate. Raw matching produces an overlap of 0.93, indicating that different target references induce almost identical selections. Mean-centering substantially reduces this average overlap, while the pairwise matrices reveal that several target pairs remain highly similar. Removing the dominant principal directions further changes the target-conditioned ranking structure while sharply reducing the matched-control correlation in the left panel.

\begin{figure*}[t]
    \centering
    \includegraphics[width=0.98\textwidth]{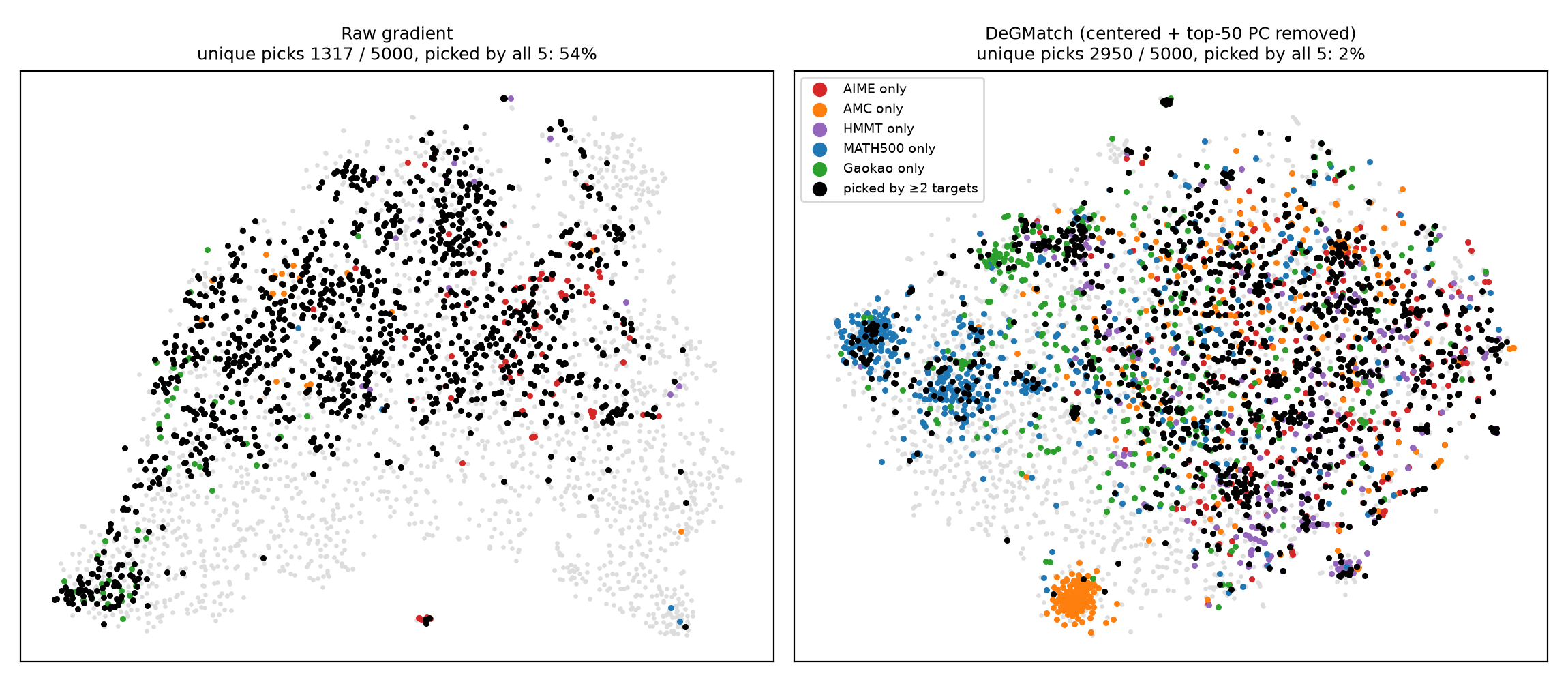}

    \caption{
    \textbf{Deconfounding reduces concentration on the same highly ranked candidates across targets.}
    t-SNE visualization of the top-1K candidates selected for each of five mathematical targets under raw-gradient matching (left) and after mean-centering and removal of the top 50 candidate principal directions (right).
    Black points denote candidates selected by at least two targets.
    Raw matching yields a union of only 1,317 distinct candidates across the five top-1K subsets, of which 54.0\% are selected by all five targets. After deconfounding, the union expands to 2,950 distinct candidates, of which only 2.4\% are selected by all five targets.
    The reported selection statistics are computed from the original selected sets rather than from the two-dimensional projection.
    }
    \label{fig:selection_visualization}
    \vspace{-10pt}
\end{figure*}

The last matrix in Figure~\ref{fig:deconf_target_sensitivity} additionally shows the final DeGMatch subsets when each target uses its own relevance gate. Their overlap is lower still, illustrating that the two stages provide complementary sources of target differentiation: semantic retrieval first restricts selection to a target-relevant candidate region, while deconfounded gradient matching makes the subsequent reranking within that region more responsive to the supplied target. This distinction also separates the effect of gradient deconfounding from the additional target specificity contributed by the relevance gate.

Importantly, this effect is not equivalent to removing trajectory-length information. The final DeGMatch score remains correlated with trajectory length ($\rho=0.69$ at $k=50$), and the magnitude term explicitly preserves information about update strength. The role of deconfounding is therefore narrower: reducing target-insensitive population structure that dominates matching while retaining other signals that can remain useful for selection.

\subsection{Target Structure in Rankings Induced by Reference Subsets}
\label{app:reference_subset_structure}

The analyses above show that removing dominant
candidate-population structure makes rankings more responsive
to changes in the supplied target references.
We next ask whether this increased sensitivity reveals
target-dependent structure rather than arbitrary dispersion.

For each of the five mathematical targets, we partition the
reference set into disjoint 10-example subsets and independently
compute the candidate ranking induced by each subset.
We repeat the partitioning multiple times and score the same
fixed 30K AIME relevance gate in every case, so changes in the
ranking geometry cannot be attributed to changes in the
candidate population.
Distances between two induced rankings are defined as one minus
their Spearman correlation.

Figure~\ref{fig:reference_subset_structure} decomposes the
effect of deconfounding step by step.
Under raw gradients, target identity can be recovered from an
induced ranking with 56\% accuracy, compared with a 20\%
five-way chance level.
Population centering alone changes this only marginally to
57\%.
Likewise, retaining only the top 50 principal components of
the centered gradients yields 56\% target identification.
In contrast, removing the population mean and these dominant
components increases target identification to 71\%.

For target identification, each reference-subset query is
assigned to the target whose other disjoint reference subsets
have the highest average Spearman correlation with its induced
ranking.
The identification score is therefore measured directly in
the original candidate-ranking space, rather than from the
two-dimensional visualization.

\begin{table}[t]
\centering
\footnotesize
\setlength{\tabcolsep}{3pt}
\renewcommand{\arraystretch}{1.06}

\begin{tabularx}{\columnwidth}{
@{}
*{5}{>{\centering\arraybackslash}X}
@{}
}
\toprule
\textbf{$k$}
& \textbf{Var.}
& \textbf{Ctrl.\ $\rho$}
& \textbf{Overlap}
& \textbf{Tokens} \\
\midrule

10
& 0.794 & 0.39 & 0.63 & 96.8M \\

25
& 0.845 & 0.36 & 0.69 & 96.0M \\

50
& 0.873 & 0.28 & 1.00 & 98.3M \\

100
& 0.897 & 0.24 & 0.73 & 98.7M \\

200
& 0.917 & 0.20 & 0.67 & 99.1M \\

\bottomrule
\end{tabularx}

\caption{
Diagnostics as the number of removed principal components varies.
\emph{Var.} is cumulative explained variance;
\emph{Ctrl.\ $\rho$} is the matched-control correlation from
Table~\ref{tab:target_specificity_deconf};
\emph{Overlap} is measured against the default selection with $k=50$.
}
\label{tab:pc_count_diagnostics}
% \vspace{-10pt}
\end{table}

The close behavior of the centered and dominant-subspace
representations further indicates that mean removal alone does
not eliminate the dominant population variation.
Across the reference-subset queries, rankings produced after
centering and rankings produced using only the top-50
principal-component subspace have Spearman correlations of
0.997--0.999, with 94--96\% overlap among their top-3K
candidates.
Thus, most of the target-conditioned ranking geometry that
remains after centering is still carried by the dominant
candidate directions.

\begin{figure}[t]
    \centering
    \includegraphics[width=\columnwidth]
    {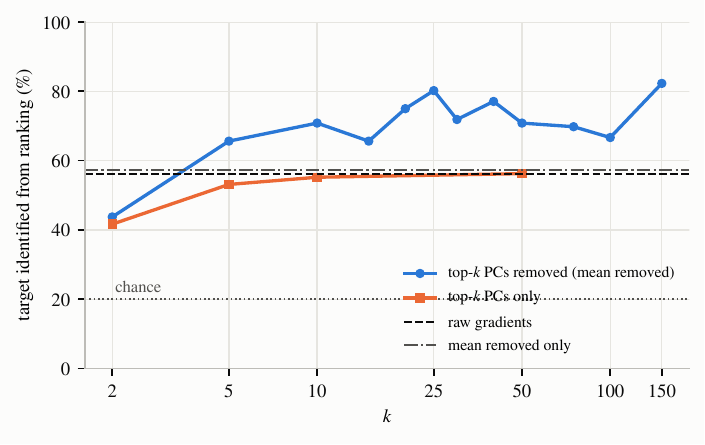}
    \caption{
    \textbf{Target-ranking structure is robust to the number
    of removed principal components.}
    Target identification is computed from candidate rankings
    induced by disjoint reference subsets, following
    Figure~\ref{fig:reference_subset_structure}.
    Blue removes the leading $k$ components after population
    centering; orange retains only those components.
    Dashed horizontal lines denote raw-gradient and
    mean-centered baselines.
    }
    \label{fig:target_ident_k}
    \vspace{-10pt}
\end{figure}

Importantly, the dominant subspace is not devoid of target
information: it retains coarse relationships among the
mathematical targets.
The relevant observation is instead that suppressing this
high-energy population structure makes finer target identity
more recoverable from the resulting candidate ordering.
This supports the interpretation that dominant shared
directions can obscure, rather than completely eliminate,
target-dependent variation.

\subsection{Visualizing Target-Conditioned Selection}
\label{app:selection_visualization}

We further visualize how deconfounding changes which candidates are highly ranked across targets. Figure~\ref{fig:selection_visualization} plots the top-1K candidates selected for each of the five mathematical targets under raw-gradient matching and after deconfounding.

Raw-gradient matching concentrates the target-conditioned selections on a small common set. Across the five top-1K lists, the union contains only 1,317 distinct candidates. Among these distinct candidates, 54.0\% are selected by all five targets and 74.1\% are selected by at least three targets. After deconfounding, the union expands to 2,950 distinct candidates; among these, only 2.4\% are selected by all five targets and 17.4\% by at least three. These statistics directly quantify the reduction in repeated selection of the same candidates across targets.

Together with the controlled ranking diagnostics above, this visualization shows the same phenomenon at the level of selected candidates: raw gradient matching repeatedly favors a common set across targets, whereas deconfounding produces substantially more target-dependent selections.

\subsection{Number of Removed Components}
\label{app:pc_sensitivity}

We use $k=50$ principal directions in the main experiments. Table~\ref{tab:pc_count_diagnostics} summarizes how the representation-level diagnostics change as $k$ varies. Increasing $k$ steadily reduces correlation with the matched non-target control, with the largest change occurring within the first several dozen removed directions. The total number of training tokens in the resulting 10K subsets remains comparatively stable across these settings.

Figure~\ref{fig:target_ident_k} provides a complementary
ranking-level view of this sensitivity.
The increase in target discrimination is not specific to the
default choice of $k=50$.
Across $k=5$--$150$, removing the dominant components yields
target-identification accuracy between 66\% and 82\%, compared
with 56\% for raw gradients and 57\% after centering alone.
In contrast, rankings formed using only the corresponding
leading components remain substantially less
target-discriminative over the evaluated values.
Removing only two components is insufficient and reduces
identification to 44\%, suggesting that the relevant shared
structure extends beyond the very first directions.

Together with the matched-control diagnostic in
Table~\ref{tab:pc_count_diagnostics}, these results indicate that
the qualitative effect of dominant-component removal is robust
over a broad range of $k$ rather than arising from a narrowly
tuned choice.
We retain $k=50$ as the fixed operating point because it lies
within this stable regime and gives the strongest downstream
average among the evaluated settings.

\begin{table}[t]
\centering
\small
\setlength{\tabcolsep}{5pt}
\renewcommand{\arraystretch}{1.06}

\begin{tabularx}{\columnwidth}{
@{}
>{\centering\arraybackslash}p{0.22\columnwidth}
*{4}{>{\centering\arraybackslash}X}
@{}
}
\toprule
\textbf{$k$}
& \textbf{AIME24}
& \textbf{AIME25}
& \textbf{AIME26}
& \textbf{Avg.} \\
\midrule

0
& 33.3 & 33.1 & 35.6 & 34.00 \\

10
& 37.1 & 32.0 & 33.0 & 34.03 \\

25
& 36.0 & 32.2 & 34.2 & 34.13 \\

\textbf{50}
& 37.9 & 31.0 & 35.6 & \textbf{34.83} \\

100
& 35.0 & 33.0 & 34.6 & 34.20 \\

\bottomrule
\end{tabularx}

\caption{
Downstream sensitivity to the number of removed principal components under a fixed 3K selection budget.
$k=0$ corresponds to population centering without principal-component removal.
Avg.\ is the unweighted mean over AIME24, AIME25, and AIME26.
}
\label{tab:pc_count_downstream}
% \vspace{-10pt}
\end{table}

The matched-control correlation decreases from 0.39 at $k=10$ to 0.28 at $k=50$, after which the improvement becomes more gradual. Meanwhile, selected-token volume remains within 96--99M tokens across all settings. We therefore use $k=50$ as a fixed operating point: the largest changes in target sensitivity occur within the first several dozen components, while further removal produces progressively smaller changes and the selected-token volume remains stable.

\begin{table}[t]
\centering
\small
\setlength{\tabcolsep}{4pt}
\renewcommand{\arraystretch}{1.06}

\begin{tabularx}{\columnwidth}{
@{}
>{\raggedright\arraybackslash}X
>{\centering\arraybackslash}p{0.38\columnwidth}
@{}
}
\toprule
\textbf{Estimator}
& \textbf{\mbox{Overlap w/ default}} \\
\midrule

Full 30K gate PCA
& 1.00 \\

6K gate sample, seed 0
& 0.87 \\

6K gate sample, seed 1
& 0.89 \\

6K gate sample, seed 2
& 0.90 \\

6K gate sample, seed 3
& 0.90 \\

Full-pool PCA
& 0.88 \\

\bottomrule
\end{tabularx}

\caption{
Sensitivity to the population used to estimate the dominant gradient subspace.
Overlap is measured against the default selection obtained from PCA over the complete 30K relevance gate.
}
\label{tab:subspace_estimation}
\vspace{-10pt}
\end{table}

\paragraph{Downstream sensitivity to the number of removed components.}
The representation-level diagnostics above show that suppressing additional dominant components progressively reduces sensitivity to the matched non-target control. We additionally test whether the choice of $k$ materially affects downstream performance. Using the fixed 3K analysis budget and otherwise identical selection and training settings, we vary the number of removed principal components and evaluate the resulting models on AIME24, AIME25, and AIME26. Table~\ref{tab:pc_count_downstream} reports the results, where $k=0$ denotes population centering without principal-component removal.

\begin{table}[t]
\centering
\footnotesize
\setlength{\tabcolsep}{2.5pt}
\renewcommand{\arraystretch}{1.06}

\begin{tabularx}{\columnwidth}{
@{}
>{\raggedright\arraybackslash}X
r
r
r
r
@{}
}
\toprule
\textbf{Target}
& \textbf{Gate}
& \textbf{$d_i<0$}
& \textbf{Pos./Budget}
& \textbf{Min.\ sel.\ $d_i$} \\
\midrule

AIME
& 30K & 0.16\% & 29,951/10K & 0.017 \\

AMC
& 30K & 0.03\% & 29,991/10K & 0.014 \\

HMMT
& 30K & 0.13\% & 29,962/10K & 0.025 \\

MATH500
& 30K & 0.04\% & 29,987/10K & 0.023 \\

Gaokao
& 30K & 0.06\% & 29,983/10K & 0.021 \\

GPQA
& 15K & 0.15\% & 14,978/5K & 0.015 \\

JEE
& 15K & 0.32\% & 14,952/5K & 0.017 \\

LiveCodeBench
& 9K & 0.24\% & 8,978/3K & 0.023 \\

CodeForces
& 9K & 1.37\% & 8,877/3K & 0.017 \\

\bottomrule
\end{tabularx}

\caption{
Directional-alignment statistics within each relevance gate.
\emph{Pos./Budget} reports the number of candidates with $d_i>0$ relative to the final selection budget, and \emph{Min.\ sel.\ $d_i$} is the minimum directional alignment among selected examples.
}
\label{tab:alignment_sign}
\vspace{-10pt}
\end{table}

Downstream performance is relatively stable across a broad range of component-removal settings, while removing the top 50 components achieves the strongest average performance among the evaluated values of $k$. In particular, the average improves from 34.00 with centering alone to 34.83 at $k=50$, while both smaller and larger removal sizes remain competitive. Together with the representation-level diagnostics above, these results support $k=50$ as a fixed operating point rather than a target-specific tuned hyperparameter. We use the same value across all targets and domains.

\begin{table*}[t]
\centering
\small
\setlength{\tabcolsep}{4.5pt}
\renewcommand{\arraystretch}{1.06}

\begin{tabularx}{\textwidth}{
@{}
>{\raggedright\arraybackslash}p{0.13\textwidth}
Y
>{\centering\arraybackslash}p{0.20\textwidth}
*{4}{Y}
@{}
}
\toprule
\textbf{Gate}
& \textbf{Size}
& \textbf{\mbox{Overlap w/ default}}
& \textbf{Tokens}
& \textbf{Median len.}
& \textbf{Median NLL}
& \textbf{AIME sim.} \\
\midrule

$1\times B$
& 10K & 0.36 & 68.9M & 5,968 & 1.26 & 0.531 \\

$2\times B$
& 20K & 0.69 & 90.0M & 8,326 & 1.49 & 0.525 \\

\textbf{$3\times B$}
& \textbf{30K} & \textbf{1.00} & \textbf{98.3M}
& \textbf{9,196} & \textbf{1.56} & \textbf{0.519} \\

$5\times B$
& 50K & 0.64 & 107.4M & 10,378 & 1.61 & 0.507 \\

No gate
& 93,733 & 0.47 & 115.8M & 11,642 & 1.66 & 0.493 \\

\bottomrule
\end{tabularx}

\caption{
Effect of relevance-gate size on the final AIME subset.
\emph{Overlap w/ default} is the fraction shared with the subset selected using the default $3B$ gate.
}
\label{tab:gate_size_diagnostics}
\vspace{-10pt}
\end{table*}

\subsection{Estimating Population Structure Within the Relevance Gate}
\label{app:subspace_estimation}

Our default implementation estimates the mean and principal directions from all examples inside the relevance gate. Table~\ref{tab:subspace_estimation} compares this choice with subsampled and full-pool alternatives.

Estimating the subspace from 6K randomly sampled gate examples changes approximately 10--13\% of the selected subset, while estimating it from the full mathematical pool produces a subset with 0.88 overlap with the default. DeGMatch defines the dominant population structure relative to the candidate population being reranked. Estimating this structure from a small random subset introduces additional sampling variability, whereas estimating it from the full pool incorporates variation outside the target-relevant candidate region. We therefore estimate the mean and principal directions from the complete relevance gate.

\subsection{Directional-Alignment Statistics}
\label{app:alignment_sign}

Finally, we examine a potential edge case of the multiplicative DeGMatch score. The directional term $d_i$ in Eq.~\ref{eq:directional_alignment} is an average cosine similarity and can therefore be negative. Because DeGMatch uses the unclamped score $s_i=d_i m_i$, a sufficiently large magnitude could in principle change the ordering among negatively aligned candidates. Table~\ref{tab:alignment_sign} reports the relevant statistics for every target used in our main experiments.

At least 98.6\% of gated candidates have positive directional alignment in every setting, and the number of positively aligned candidates substantially exceeds the selection budget. More importantly, every trajectory selected by DeGMatch has $d_i>0$, with the minimum selected value ranging from 0.014 to 0.025.

As a direct check, replacing the implemented score with
$s_i=\max(d_i,0)m_i$ produces exactly the same selected subset for every target. Thus, although the unclamped product admits a theoretical ordering ambiguity among negatively aligned candidates, negative alignment does not affect any selection or downstream result reported in this work.

\begin{table}[t]
\centering
\small
\setlength{\tabcolsep}{5pt}
\renewcommand{\arraystretch}{1.06}

\begin{tabularx}{\columnwidth}{
@{}
>{\centering\arraybackslash}p{0.24\columnwidth}
*{4}{>{\centering\arraybackslash}X}
@{}
}
\toprule
\textbf{Gate size}
& \textbf{AIME24}
& \textbf{AIME25}
& \textbf{AIME26}
& \textbf{Avg.} \\
\midrule

9K
& 34.6 & 32.3 & 36.2 & 34.37 \\

15K
& 37.5 & 31.5 & 34.0 & 34.33 \\

\textbf{30K}
& 37.9 & 31.0 & 35.6 & \textbf{34.83} \\

Full pool
& 37.3 & 29.7 & 34.7 & 33.90 \\

\bottomrule
\end{tabularx}

\caption{
Downstream sensitivity to the relevance-gate size under a fixed 3K selection budget.
\emph{Full pool} removes the relevance restriction and allows gradient reranking over the complete mathematical candidate pool.
Avg.\ is the unweighted mean over AIME24, AIME25, and AIME26.
}
\label{tab:gate_size_downstream}
\vspace{-10pt}
\end{table}

\section{Analysis of the Relevance Gate}
\label{app:gate_analysis}

We analyze how the relevance gate shapes the candidate region before gradient-based reranking. This section examines the first stage of DeGMatch and verifies that the relevance gate restricts gradient-based selection to a target-relevant candidate region without determining the final selected subset. Unless otherwise specified, the analyses use the AIME target with a 10K final selection budget. The subset-level diagnostics characterize how the gate changes the candidate region and selected data, while Section~\ref{app:gate_size} additionally evaluates downstream sensitivity to gate size.

\subsection{Effect of Gate Size}
\label{app:gate_size}

The gate size controls how far gradient-based selection can search beyond the trajectories most similar to the target references. Table~\ref{tab:gate_size_diagnostics} varies the gate size while keeping the final selection budget fixed at $B=10$K.

A smaller gate constrains selection to trajectories that are more semantically similar to the target references, while expanding the gate gives gradient matching access to longer and higher-NLL trajectories. When $G=B$, the gate itself determines the final subset, reducing selection to semantic retrieval. At the other extreme, removing the gate increases the selected token volume from 68.9M to 115.8M and reduces AIME similarity from 0.531 to 0.493.

The default $3B$ gate lies between these two extremes. It retains a target-relevant candidate region while leaving two-thirds of the gated examples available for student-specific reranking. These subset-level statistics alone do not establish a downstream-optimal gate size; we therefore evaluate downstream sensitivity separately below.

\begin{table*}[t]
\centering
\small
\setlength{\tabcolsep}{4.5pt}
\renewcommand{\arraystretch}{1.06}

\begin{tabularx}{\textwidth}{
@{}
>{\raggedright\arraybackslash}p{0.29\textwidth}
c
c
c
c
@{}
}
\toprule
\textbf{Gate construction}
& \textbf{Gate Jaccard}
& \textbf{Top-1K coverage min./med.}
& \textbf{Selection overlap}
& \textbf{AIME sim.} \\
\midrule

\textbf{Qwen3 + round-robin}
& \textbf{1.00}
& \textbf{893 / 949}
& \textbf{1.00}
& 0.519 \\

Student hidden state + round-robin
& 0.30
& 214 / 506
& 0.47
& 0.499 \\

Qwen3 + max similarity
& 0.68
& 96 / 913
& 0.76
& \textbf{0.522} \\

Qwen3 + mean similarity
& 0.40
& 40 / 702
& 0.56
& 0.515 \\

\bottomrule
\end{tabularx}

\caption{
Effect of the embedding representation and retrieval aggregation rule.
\emph{Top-1K coverage} reports, across target references, the minimum and median number of each reference's 1,000 nearest candidates retained in the 30K gate.
\emph{Selection overlap} is measured against the final subset obtained with the default gate.
}
\label{tab:gate_construction}
\vspace{-10pt}
\end{table*}

\paragraph{Downstream sensitivity to gate size.}
The preceding diagnostics characterize how the relevance gate changes the candidate region and the resulting selected subset. We additionally evaluate whether these differences translate to downstream performance. Using a fixed 3K final selection budget, we vary the AIME relevance-gate size while keeping the gradient representation, scoring rule, and downstream training procedure unchanged. Table~\ref{tab:gate_size_downstream} reports the resulting downstream performance.

Performance remains similar across the finite gate sizes, with the 30K gate achieving the strongest average among the evaluated settings. In contrast, removing the relevance restriction and reranking over the full candidate pool reduces the average score to 33.90. These results suggest that the benefit of the relevance stage is not tied to a narrowly tuned gate size: a range of target-restricted candidate regions performs comparably, while eliminating the gate altogether is less effective.

\subsection{Embedding Representation and Retrieval Aggregation}
\label{app:gate_construction_analysis}

The round-robin construction in Section~\ref{sec:relevance_gate} is designed to prevent references with unusually dense neighborhoods from dominating the relevance gate. Table~\ref{tab:gate_construction} compares the default construction with alternative embedding representations and aggregation rules, while fixing the gate size at 30K.

Round-robin retrieval provides substantially more balanced coverage across target references. Under the default construction, even the least-covered reference retains 893 of its 1,000 nearest candidates, whereas the minimum coverage falls to 96 under max-similarity aggregation and to 40 under mean-similarity aggregation. Notably, max-similarity aggregation achieves slightly higher overall AIME similarity than round-robin (0.522 vs.\ 0.519) despite this sharp reduction in worst-case coverage. This supports round-robin retrieval as a coverage-preserving rather than similarity-maximizing design, preventing the gate from concentrating around only a subset of reference neighborhoods.

The choice of embedding representation also matters. Replacing Qwen3-Embedding-8B with hidden states from the initial student yields a gate with a Jaccard similarity of 0.30 to the default gate and reduces both neighborhood coverage and AIME similarity. We therefore use a dedicated embedding model to identify the semantically relevant candidate region and reserve the student model for the subsequent gradient-based utility estimate.

\subsection{Post-Hoc Proximity to Evaluation Problems}
\label{app:test_proximity}

We additionally examine whether the target-reference gate identifies problems related to held-out evaluation instances. This analysis is strictly post hoc: evaluation problems are never used to construct the gate, compute gradients, or select training examples.

For each AIME24 or AIME25 problem, we retrieve its ten nearest problems from a given training set using the same Qwen3 problem embeddings and average their cosine similarities. Table~\ref{tab:test_proximity} reports this nearest-neighbor proximity together with the average number of neighbors per evaluation problem whose cosine similarity exceeds 0.7.

\begin{table}[t]
\centering
\small
\setlength{\tabcolsep}{4pt}
\renewcommand{\arraystretch}{1.06}

\begin{tabularx}{\columnwidth}{
@{}
>{\raggedright\arraybackslash}X
c
c
c
@{}
}
\toprule
\textbf{Set}
& \textbf{AIME24}
& \textbf{AIME25}
& \textbf{$\cos>0.7$} \\
\midrule

Random 10K
& 0.648 & 0.639 & 1.0 \\

Length 10K
& 0.658 & 0.642 & 1.4 \\

Relevance gate 30K
& \textbf{0.691} & \textbf{0.681} & \textbf{7.0} \\

DeGMatch 10K
& 0.668 & 0.657 & 2.9 \\

\bottomrule
\end{tabularx}

\caption{
Post-hoc semantic proximity to AIME24/25 evaluation problems.
The first two columns report the mean cosine similarity of the ten nearest training problems per evaluation problem.
The final column reports the mean number of neighbors with cosine similarity above 0.7.
Evaluation problems are used only for this diagnostic.
}
\label{tab:test_proximity}
\vspace{-10pt}
\end{table}

\begin{table*}[t]
\centering
\small
\setlength{\tabcolsep}{3.8pt}
\renewcommand{\arraystretch}{1.06}

\begin{tabularx}{\textwidth}{
@{}
>{\raggedright\arraybackslash}p{0.08\textwidth}
>{\centering\arraybackslash}p{0.19\textwidth}
>{\centering\arraybackslash}p{0.14\textwidth}
*{4}{Y}
@{}
}
\toprule
\textbf{Set}
& \textbf{Length (tokens)}
& \textbf{Student NLL}
& \textbf{Grad.\ norm}
& \textbf{AIME sim.}
& \textbf{NT+Comb.}
& \textbf{Olympiad} \\
\midrule

Pool
& 4,935 [2,959, 8,107]
& 1.14 [0.90, 1.46]
& 31.2
& 0.47
& 0.29
& 0.73 \\

Random
& 4,974 [3,019, 8,116]
& 1.15 [0.91, 1.47]
& 31.5
& 0.47
& 0.29
& 0.73 \\

Length
& 13,940 [12,669, 15,574]
& 1.77 [1.58, 2.01]
& 42.3
& 0.49
& 0.38
& 0.67 \\

MaxPPL
& 11,786 [9,066, 14,460]
& 1.95 [1.84, 2.13]
& 42.2
& 0.48
& 0.56
& 0.70 \\

RDS+
& 10,104 [7,403, 13,364]
& 1.50 [1.28, 1.74]
& 41.2
& 0.50
& 0.47
& 0.63 \\

DeGMatch
& 9,196 [6,823, 12,600]
& 1.56 [1.33, 1.82]
& 41.0
& 0.52
& 0.48
& 0.64 \\

\bottomrule
\end{tabularx}

\caption{
Properties of mathematical subsets selected for AIME.
Length and student NLL are reported as median [IQR] when available.
NT+Comb.\ is the combined fraction of examples whose dataset-provided \texttt{problem\_type} is Number Theory or Combinatorics; Olympiad is the fraction whose dataset-provided \texttt{source} is \texttt{olympiads}.
}
\label{tab:selected_properties}
\vspace{-10pt}
\end{table*}

\begin{table}[t]
\centering
\footnotesize
\setlength{\tabcolsep}{3pt}
\renewcommand{\arraystretch}{1.06}

\begin{tabularx}{\columnwidth}{
@{}
>{\raggedright\arraybackslash}p{0.20\columnwidth}
*{6}{>{\centering\arraybackslash}X}
@{}
}
\toprule
\textbf{Method}
& \textbf{Tokens}
& \textbf{Med. len.}
& \textbf{A24}
& \textbf{A25}
& \textbf{A26}
& \textbf{Avg.} \\
\midrule

Rand-LC
& 35.3M
& 12,102
& \textbf{38.4}
& 30.6
& 32.1
& 33.70 \\

\textbf{DeGMatch}
& 36.1M
& 12,274
& 37.9
& \textbf{31.0}
& \textbf{35.6}
& \textbf{34.83} \\

\bottomrule
\end{tabularx}

\caption{
Length-controlled downstream comparison under a fixed 3K selection budget.
Rand-LC (length-controlled random sampling) draws from the same 30K AIME relevance gate as DeGMatch, excluding DeGMatch-selected examples, and approximately matches its complete-sequence length distribution using 20 quantile strata.
}
\label{tab:length_controlled_comparison}
\vspace{-10pt}
\end{table}

The 30K relevance gate is substantially closer to AIME24/25 than Random or Length, providing additional evidence that the independently constructed reference set retrieves a candidate region related to the intended target capability. The final DeGMatch subset remains more target-proximal than Random and Length, but is less similar than the full relevance gate.

The lower proximity of the final DeGMatch subset relative to the full gate is consistent with the role of the second stage: after semantic retrieval defines a relevant candidate region, gradient matching is not constrained to choose the semantically nearest problems. Instead, it reranks the gate according to student-specific training utility. We emphasize that these evaluation-set similarities are used only for retrospective analysis and play no role in the selection procedure.

\section{Characteristics of Selected Reasoning Data}
\label{app:selected_data}

We examine the characteristics of the trajectories selected by DeGMatch and compare them with those produced by heuristic and target-aware alternatives. Unless otherwise specified, this section analyzes 10K mathematical trajectories selected for the AIME target.

\begin{table}[t]
\centering
\footnotesize
\setlength{\tabcolsep}{3.2pt}
\renewcommand{\arraystretch}{1.06}

\begin{tabularx}{\columnwidth}{
@{}
>{\raggedright\arraybackslash}p{0.22\columnwidth}
*{5}{Y}
@{}
}
\toprule
& \textbf{AIME}
& \textbf{AMC}
& \textbf{HMMT}
& \textbf{MATH}
& \textbf{GK} \\
\midrule

AIME
& 1.00 & 0.21 & 0.27 & 0.12 & 0.11 \\

AMC
&      & 1.00 & 0.20 & 0.13 & 0.12 \\

HMMT
&      &      & 1.00 & 0.11 & 0.12 \\

MATH500
&      &      &      & 1.00 & 0.14 \\

Gaokao
&      &      &      &      & 1.00 \\

\bottomrule
\end{tabularx}

\caption{
Pairwise Jaccard overlap among target-specific 10K DeGMatch subsets.
Two independently sampled uniform 10K subsets from the 93,733-example mathematical pool have expected Jaccard overlap of approximately 0.056.
}
\label{tab:degmatch_target_overlap}
\vspace{-10pt}
\end{table}

\subsection{Properties of the Selected Subset}
\label{app:selected_properties}

Table~\ref{tab:selected_properties} compares the resulting subsets in terms of trajectory length, student NLL, gradient norm, semantic similarity to the AIME references, and coarse problem composition. To summarize subject composition compactly, we report the combined fraction of examples labeled Number Theory or Combinatorics in the dataset metadata, together with the fraction carrying the \texttt{olympiads} source tag.

DeGMatch does not simply reproduce a single coarse selection heuristic. Its trajectories are substantially shorter than those selected by Length, with a median of 9.2K versus 13.9K tokens, while their student NLL is less extreme than MaxPPL. At the same time, DeGMatch has the highest semantic similarity to the AIME references among the compared 10K subsets.

The instance-level overlap tells a similar story. DeGMatch shares only 31\% of its examples with Length, 30\% with MaxPPL, 35\% with RDS+, and 11\% with Random. Its selected subset therefore cannot be characterized solely as choosing long trajectories, high-loss trajectories, or the examples most similar under a representation-level target metric.

\begin{table*}[t]
\centering
\footnotesize
\setlength{\tabcolsep}{4pt}
\renewcommand{\arraystretch}{1.06}
\begin{tabularx}{\textwidth}{
@{}
>{\raggedright\arraybackslash}p{0.34\textwidth}
*{7}{>{\centering\arraybackslash}X}
@{}
}
\toprule
\textbf{Rank shift}
& \textbf{\#}
& \textbf{Med.\ len.}
& \textbf{Med.\ NLL}
& \textbf{Alg.}
& \textbf{Comb.}
& \textbf{Geom.}
& \textbf{NT} \\
\midrule

Raw top 1K $\rightarrow$ below 10K
& 258
& 9,374
& 1.57
& 66
& 42
& 104
& 39 \\

Raw below 10K $\rightarrow$ DeGMatch top 1K
& 276
& 14,572
& 1.85
& 28
& 102
& 47
& 86 \\

\bottomrule
\end{tabularx}

\caption{
Candidate groups undergoing large rank changes after deconfounding within the AIME relevance gate.
The final four columns report counts of algebra, combinatorics, geometry, and number-theory problems; less frequent categories are omitted.
}
\label{tab:rank_changes}
\vspace{-10pt}
\end{table*}

\paragraph{Length-controlled comparison.}
The descriptive statistics above show that DeGMatch differs substantially from explicit length-based selection, but trajectory length could still partially explain its downstream performance. We therefore construct a length-controlled random baseline from the same 30K AIME relevance gate used by the 3K DeGMatch analysis. After excluding the DeGMatch-selected examples, we partition the remaining candidates into 20 quantiles by complete training-sequence length and sample from each stratum to match the length distribution of the DeGMatch subset. Sampling uses seed 0 and does not use gradients.\footnote{The longest stratum is short by 116 examples after excluding the DeGMatch subset; we fill this deficit from the adjacent shorter stratum.} Table~\ref{tab:length_controlled_comparison} reports the downstream comparison.

DeGMatch achieves a higher average score than the length-controlled random baseline (34.83 vs.\ 33.70), including improvements on AIME25 and AIME26. The two subsets have zero instance overlap while closely matching in both total token volume (36.1M vs.\ 35.3M) and median trajectory length (12,274 vs.\ 12,102 tokens). Because both subsets are drawn from the same relevance gate and have closely matched length distributions, the performance difference cannot be explained solely by semantic gating or trajectory length. This provides complementary evidence that gradient-based reranking captures information beyond these two factors.

\subsection{Target-Specificity of the Selected Subsets}
\label{app:selected_overlap}

We next examine how strongly target-aware selectors respond to changes in the supplied target references. For each method, we independently select 10K trajectories for AIME, AMC, Gaokao, HMMT, and MATH500 and measure the pairwise Jaccard overlap among the resulting target-specific subsets.

Table~\ref{tab:degmatch_target_overlap} first reports the full pairwise overlap for DeGMatch. Competition-style targets share more selected data with each other than with MATH500 or Gaokao. AIME--HMMT overlap is 0.27 and AIME--AMC overlap is 0.21, whereas AIME overlaps with MATH500 and Gaokao by only 0.12 and 0.11. The target-specific subsets therefore share a common pool of potentially useful reasoning trajectories while still changing substantially with the supplied target capability.

Comparisons with existing target-aware selectors are provided in Appendix~\ref{app:existing_target_sensitivity}. DeGMatch has a mean pairwise overlap of 0.15, compared with 0.253 at the relevance-gate level, showing that the gradient stage adds target differentiation beyond semantic retrieval alone. Importantly, we do not interpret lower overlap as intrinsically better: TAROT produces even lower cross-target overlap, yet does not outperform the strongest target-agnostic baselines in downstream evaluation. Rather, these diagnostics show that existing target-aware methods differ substantially in how strongly their selections respond to the supplied target references, and that DeGMatch produces target-dependent subsets without explicitly optimizing for low overlap.

\subsection{How Deconfounding Changes Candidate Rankings}
\label{app:rank_changes}

We finally examine which candidates undergo the largest rank changes after dominant-component removal. Table~\ref{tab:rank_changes} compares trajectories that move from the raw-gradient top 1K to below rank 10K with trajectories that move from outside the raw top 10K into the DeGMatch top 1K.

The promoted examples are not simply shorter or lower-loss trajectories. In fact, candidates moving into the DeGMatch top 1K are longer and have higher median student NLL than those strongly demoted by deconfounding. This provides an additional indication that dominant-component removal does not act as a simple correction for trajectory length or difficulty.

The problem-type composition also changes: the promoted group contains more combinatorics and number-theory problems, whereas the demoted group contains more geometry and algebra problems. We treat this shift as descriptive rather than as evidence that any particular subject category is intrinsically more useful. The important observation is that removing dominant population structure substantially reorders candidates even when they occupy similar ranges of coarse trajectory characteristics.

\begin{table*}[t]
\centering
\small
\setlength{\tabcolsep}{4pt}
\renewcommand{\arraystretch}{1.10}

\begin{tabularx}{\textwidth}{
@{}
>{\centering\arraybackslash}p{0.05\textwidth}
>{\raggedright\arraybackslash}X
>{\raggedright\arraybackslash}X
>{\centering\arraybackslash}p{0.08\textwidth}
>{\centering\arraybackslash}p{0.07\textwidth}
>{\centering\arraybackslash}p{0.09\textwidth}
@{}
}
\toprule
\textbf{Ex.}
& \textbf{Selected training problem}
& \textbf{Shared reasoning structure}
& \textbf{Deconf. rank}
& \textbf{Raw rank}
& \textbf{Semantic rank} \\
\midrule

A
&
Six people are seated in a row, where three designated people must not sit next to one another.
&
Forbidden-adjacency arrangement problem closely matching the reference at both the problem and solution level.
&
\textbf{1}
&
6,879
&
1
\\
\midrule
B
&
Three red, two white, and two blue balls are arranged in a row; compute the probability that the two white balls or the two blue balls are adjacent.
&
Both solutions define adjacency events, merge adjacent objects into blocks, count the resulting arrangements, and combine overlapping cases using inclusion--exclusion.
&
\textbf{13}
&
14,215
&
4,862
\\

\bottomrule
\end{tabularx}

\caption{
Illustrative target-conditioned matches for the AIME 2020-II-9 reference.
All ranks are computed among the 30K examples in the relevance gate for this reference, with rank 1 denoting the closest match.
Semantic rank uses Qwen3 problem-embedding cosine similarity; the corresponding cosine similarities are 0.80 for Example A and 0.43 for Example B.
}
\label{tab:qualitative_matches}
\vspace{-10pt}
\end{table*}

\subsection{Illustrative Target Matches}
\label{app:qualitative_matches}

We provide qualitative examples of how deconfounded gradient matching relates target references to selected training trajectories. For a given target reference, ranks are computed among the 30K examples in its AIME relevance gate. To avoid trivial matches, we exclude candidate--reference pairs whose Qwen3 problem-embedding cosine similarity is at least 0.90. This filtering is used only for the qualitative analysis.

Table~\ref{tab:qualitative_matches} presents two candidates selected for the same AIME reference, AIME 2020-II-9, a combinatorics problem involving arrangements subject to forbidden adjacencies. The first candidate is also semantically close to the reference, whereas the second is semantically more distant but uses a closely related reasoning procedure.

Example A is an intuitive match: both the reference and selected problem ask for arrangements under forbidden-adjacency constraints. Although it is the nearest candidate by semantic similarity, it ranks only 6,879th under raw-gradient cosine and rises to rank 1 after deconfounding. This example illustrates how dominant population-level gradient structure can down-rank even a semantically obvious target match under raw gradient similarity.

\begin{table}[t]
\centering
\footnotesize
\setlength{\tabcolsep}{2pt}
\renewcommand{\arraystretch}{1.12}
\renewcommand{\tabularxcolumn}[1]{m{#1}}

\begin{tabularx}{\columnwidth}{
@{}
>{\raggedright\arraybackslash}X
>{\centering\arraybackslash}c
>{\centering\arraybackslash}X
>{\centering\arraybackslash}c
>{\centering\arraybackslash}c
@{}
}
\toprule
\textbf{Variant}
& \textbf{Rel.}
& \textbf{Transform}
& \textbf{Dir.}
& \textbf{Mag.} \\
\midrule

Random
& -- & -- & -- & -- \\

Relevance only
& \checkmark & -- & -- & -- \\

Raw-gradient dir.
& \checkmark & None & \checkmark & -- \\

Centered dir.
& \checkmark & Center & \checkmark & -- \\

Deconfounded dir.
& \checkmark & Center + PC50 & \checkmark & -- \\

Deconfounded mag.
& \checkmark & Center + PC50 & -- & \checkmark \\

\textbf{DeGMatch}
& \checkmark & Center + PC50 & \checkmark & \checkmark \\

\bottomrule
\end{tabularx}

\caption{
Component-validation variants.
Rel. denotes target-relevance filtering; Center denotes candidate-population mean subtraction; PC50 denotes removal of the top 50 principal components; and Dir. and Mag. denote directional alignment and target-subspace magnitude, respectively.
}
\label{tab:component_variants}
\vspace{-10pt}
\end{table}

\begin{table*}[t]
\centering
\footnotesize
\setlength{\tabcolsep}{3pt}
\renewcommand{\arraystretch}{1.12}

\begin{tabularx}{\textwidth}{
@{}
>{\hsize=2\hsize\raggedright\arraybackslash}X
*{10}{>{\hsize=.9\hsize\centering\arraybackslash}X}
@{}
}
\toprule
\textbf{Method}
& \textbf{AIME24}
& \textbf{AIME25}
& \textbf{AIME26}
& \textbf{AMC23}
& \textbf{HMMT}
& \textbf{MATH500}
& \textbf{GK24}
& \textbf{GK25}
& \textbf{Avg.}
& \textbf{Tokens} \\
\midrule

Length
& 45.0 & 33.3 & 40.4 & 82.3 & 21.7 & 91.8 & 74.3 & 81.0
& 58.73 & 142.3M \\

Gate30K+Length
& 45.5 & 35.0 & 40.0 & 83.1 & 23.0 & 92.0 & 74.3 & 80.0
& 59.11 & 114.6M \\

MaxPPL
& 41.7 & 34.6 & 38.5 & 82.0 & 20.3 & 92.4 & 73.1 & 80.8
& 57.93 & 117.2M \\

Gate30K+MaxPPL
& 44.8 & 31.7 & 41.9 & 82.7 & 22.0 & 92.8 & 73.6 & 80.8
& 58.79 & 105.5M \\

\textbf{DeGMatch}
& 47.7 & 33.8 & 42.6 & 85.6 & 23.7 & 92.8 & 76.9 & 82.7
& \textbf{60.73} & 86.9M \\

\bottomrule
\end{tabularx}

\caption{
Gate-controlled heuristic baselines under a fixed 10K-example budget.
\emph{Gate30K} applies the corresponding heuristic only within the same 30K AIME relevance gate used by DeGMatch.
Avg.\ is the unweighted mean over the eight mathematical benchmarks.
}
\label{tab:gate_heuristic_controls}
\vspace{-10pt}
\end{table*}

Example B illustrates a different behavior. Its surface form is substantially less similar to the reference, ranking 4,862th by problem-embedding similarity and 14,215th by raw-gradient cosine, but it rises to rank 13 after deconfounding. Inspection of the reasoning trajectories reveals a shared solution procedure: both define forbidden-adjacency events, merge adjacent objects into single blocks to count constrained arrangements, and use inclusion--exclusion to combine overlapping cases. The match is therefore consistent with DeGMatch prioritizing candidates with shared reasoning structure despite limited surface similarity.

These examples are intended only as qualitative illustrations. The aggregate analyses in Sections~\ref{app:deconf_target_specificity} and~\ref{app:rank_changes} provide quantitative evidence for the broader effect of deconfounding on target-conditioned rankings.

\section{Additional Experimental Results}
\label{app:additional_results}

\subsection{Component-Validation Details}
\label{app:component_details}

Table~\ref{tab:component_variants} summarizes the variants used in Section~\ref{sec:component_ablation}. All gradient-based variants select 3K examples from the same fixed 30K AIME relevance gate and differ only in the gradient transformation or scoring terms used within the gate. Relevance only selects directly from the semantic retrieval ranking, while Random samples from the full candidate pool. Raw-gradient direction uses directional target matching without gradient transformation; Centered direction additionally subtracts the candidate-population mean; and Deconfounded direction further removes the top 50 principal directions. Deconfounded magnitude uses the same transformed gradients but scores candidates only by target-subspace magnitude, whereas DeGMatch combines directional alignment and target-subspace magnitude multiplicatively.

\begin{table}[t]
\centering
\footnotesize
\setlength{\tabcolsep}{3pt}
\renewcommand{\arraystretch}{1.08}

\resizebox{\columnwidth}{!}{%
\begin{tabular}{
@{}
l
l
rrrr
r
@{}
}
\toprule
\textbf{Variant}
& \textbf{Score}
& \textbf{A24}
& \textbf{A25}
& \textbf{A26}
& \textbf{Avg.}
& \textbf{Tokens} \\
\midrule

Direction only ($\gamma=0$)
& $d_i$
& 34.6 & 31.9 & 36.7 & 34.40 & 30.8M \\

Weak magnitude ($\gamma=\frac{1}{2}$)
& $d_i\sqrt{m_i}$
& 36.0 & 30.0 & 34.6 & 33.53 & 34.6M \\

\textbf{DeGMatch ($\gamma=1$)}
& $\mathbf{d_i m_i}$
& \textbf{37.9} & 31.0 & \textbf{35.6}
& \textbf{34.83} & 36.1M \\

Strong magnitude ($\gamma=2$)
& $d_i m_i^2$
& 37.1 & 32.1 & 34.6 & 34.60 & 37.1M \\

\midrule

Length-normalized
& $d_i m_i/\sqrt{T_i}$
& 34.2 & 30.0 & 32.1 & 32.10 & 28.1M \\

Length-residualized
& $d_i\exp(m_i^{\mathrm{res}})$
& 36.0 & \textbf{32.9} & 35.0 & 34.63 & 27.1M \\

\bottomrule
\end{tabular}%
}

\caption{
Sensitivity to alternative magnitude scalings under a fixed 3K selection budget.
$T_i$ denotes trajectory length and $m_i^{\mathrm{res}}$ denotes the length-residualized magnitude.
All variants use the same deconfounded gradient representation and 30K AIME relevance gate.
}
\label{tab:magnitude_scaling}
\vspace{-10pt}
\end{table}

\paragraph{Gate-controlled heuristic baselines.}
The relevance gate itself may improve strong target-agnostic heuristics by restricting selection to a target-relevant candidate region. To separate this effect from the subsequent gradient-based reranking, we apply Length and MaxPPL directly within the same 30K AIME relevance gate used by DeGMatch. Table~\ref{tab:gate_heuristic_controls} reports the resulting 10K-example subsets.

Relevance gating improves both heuristics: Length increases from 58.73 to 59.11 average performance, while MaxPPL increases from 57.93 to 58.79. However, DeGMatch remains stronger at 60.73 while using fewer training tokens. Thus, the performance of DeGMatch is not explained by simply applying a strong length- or difficulty-based heuristic within the relevance gate.

\paragraph{Sensitivity to magnitude scaling.}
We further examine the role of target-subspace magnitude in the DeGMatch score.
Prior work has shown that sequence length can systematically affect
gradient-based influence scores, motivating gradient normalization to mitigate
length-dependent selection bias~\citep{wang2025dynamic}.
We therefore test both how strongly magnitude should enter the score and
whether its contribution can be explained primarily by trajectory length.

We first vary the strength of the magnitude term using
$d_i m_i^\gamma$ with $\gamma\in\{0,\frac{1}{2},1,2\}$, where
$\gamma=0$ corresponds to directional matching alone and $\gamma=1$ is the
default DeGMatch score.
In our setting, $m_i$ is strongly positively correlated with trajectory length
(Spearman $\rho=0.85$), so we additionally consider two length-adjusted
variants.
The first uses $d_i m_i/\sqrt{T_i}$, where $T_i$ is the number of tokens in the
complete training sequence.
For the second, we fit
\[
\log m_i=\hat{\alpha}+\hat{\beta}\log T_i+\epsilon_i
\]
by ordinary least squares over all 30K candidates in the AIME relevance gate
and define
\[
m_i^{\mathrm{res}}
=
\log m_i-(\hat{\alpha}+\hat{\beta}\log T_i).
\]
We obtain $\hat{\beta}=0.54$.
The resulting score $d_i\exp(m_i^{\mathrm{res}})$ is rank-equivalent, up to a
positive constant, to $d_i m_i/T_i^{\hat{\beta}}$ and reduces the Spearman
correlation with trajectory length from $0.85$ to $0.08$.

Table~\ref{tab:magnitude_scaling} reports the downstream results.
The default $d_i m_i$ achieves the strongest average performance,
while the squared-magnitude and length-residualized variants
remain close.
Notably, removing most of the empirical association between magnitude and
trajectory length leaves performance nearly unchanged, suggesting that the
benefit of target-subspace magnitude is not explained solely by selecting
longer trajectories.
Overall, the results support the simple multiplicative score while indicating
that its effectiveness is not narrowly tied to either a particular magnitude
exponent or trajectory length.

\subsection{Per-Target Selection Statistics}
\label{app:selection_statistics}

\begin{table*}[t]
\centering
\small
\setlength{\tabcolsep}{4.5pt}
\renewcommand{\arraystretch}{1.08}
\resizebox{\textwidth}{!}{
\begin{tabularx}{\textwidth}{
@{}
>{\raggedright\arraybackslash}p{0.14\textwidth}
*{6}{Y}
@{}
}
\toprule
\textbf{Method}
& \textbf{AIME}
& \textbf{AMC}
& \textbf{Gaokao}
& \textbf{MATH500}
& \textbf{HMMT}
& \textbf{Avg.} \\
\midrule

\multicolumn{7}{l}{\textit{Heuristic}} \\

Random
& \multicolumn{5}{c}{60.3 (6.0K)}
& 60.3 \\

Length
& \multicolumn{5}{c}{142.3 (14.2K)}
& 142.3 \\

MaxPPL
& \multicolumn{5}{c}{117.2 (11.7K)}
& 117.2 \\

\midrule
\multicolumn{7}{l}{\textit{Target-aware}} \\

RDS+
& 104.9 (10.5K)
& 91.0 (9.1K)
& 63.3 (6.3K)
& 56.6 (5.7K)
& 110.0 (11.0K)
& 85.2 \\

LESS
& 76.8 (7.7K)
& 72.7 (7.3K)
& 64.6 (6.5K)
& 61.3 (6.1K)
& 77.8 (7.8K)
& 70.6 \\

GIST
& 85.1 (8.5K)
& 87.4 (8.7K)
& 87.7 (8.8K)
& 85.1 (8.5K)
& 85.4 (8.5K)
& 86.1 \\

TAROT
& 80.3 (8.0K)
& 75.0 (7.5K)
& 68.9 (6.9K)
& 67.6 (6.8K)
& 81.6 (8.2K)
& 74.7 \\

\midrule
\multicolumn{7}{l}{\textit{Reasoning-specific}} \\

RSR
& \multicolumn{5}{c}{34.9 (3.5K)}
& 34.9 \\

TEMP
& \multicolumn{5}{c}{64.9 (6.5K)}
& 64.9 \\
\midrule
{DeGMatch}
& {98.3 (9.8K)}
& {90.8 (9.1K)}
& {73.8 (7.4K)}
& {69.7 (7.0K)}
& {102.0 (10.2K)}
& {86.9} \\

\bottomrule
\end{tabularx}
}
\caption{
Training tokens for each 10K-example mathematical subset, in millions, with mean tokens per example in parentheses. Target-agnostic methods use the same subset for every target, whereas target-aware methods select a separate subset for each target. No selected sequence exceeds the 32,768-token training limit, so no example is truncated.
}
\label{tab:target_tokens}
\vspace{-10pt}
\end{table*}

Because reasoning trajectories vary substantially in length, equal example budgets can correspond to very different amounts of training supervision. Table~\ref{tab:target_tokens} reports the number of training tokens in each 10K-example mathematical subset. Counts are measured with the student-model tokenizer over the complete training sequence and correspond to a single training epoch. Select2Reason is omitted because we evaluate its released checkpoint rather than a budget-matched subset selected under our pipeline.

Despite using the same 10K-example budget, the target-agnostic subsets vary substantially in training-token volume, from 34.9M tokens for RSR to 142.3M for Length, a 4.1$\times$ difference. In particular, RSR contains substantially fewer training tokens than Random (34.9M vs.\ 60.3M), illustrating that matching the number of selected examples does not imply a matched amount of training supervision. Target-aware methods also vary across target capabilities: for example, RDS+ ranges from 56.6M tokens for MATH500 to 110.0M for HMMT, while DeGMatch ranges from 69.7M to 102.0M. These statistics motivate reporting both example and token budgets when comparing reasoning-data selection methods.

\begin{table}[t]
\centering
\small
\setlength{\tabcolsep}{4pt}
\renewcommand{\arraystretch}{1.10}

\resizebox{\columnwidth}{!}{%
\begin{tabular}{
@{}
l
r
rrrr
r
@{}
}
\toprule
\textbf{Method} 
& \textbf{\# Ex.}
& \textbf{AIME24}
& \textbf{AIME25}
& \textbf{AIME26}
& \textbf{Avg.}
& \textbf{Tokens}
\\
\midrule

\multirow{5}{*}{Random}
& 1K  & 27.7 & 22.9 & 29.2 & 26.60 & 6.1M  \\
& 3K  & 30.4 & 27.3 & 32.3 & 30.00 & 18.0M \\
& 5K  & 39.4 & 30.0 & 30.8 & 33.40 & 30.1M \\
& 10K & 39.2 & 30.8 & 34.4 & 34.80 & 60.3M \\
& 15K & 41.2 & 30.8 & 39.2 & 37.07 & 90.0M \\
\midrule

\multirow{5}{*}{Length}
& 1K  & 30.8 & 24.8 & 29.4 & 28.33 & 17.7M  \\
& 3K  & 36.9 & 30.4 & 36.3 & 34.53 & 49.7M  \\
& 5K  & 39.4 & 31.7 & 37.9 & 36.33 & 78.8M  \\
& 10K & 45.0 & 33.3 & 40.4 & 39.57 & 142.3M \\
& 15K & 43.5 & 33.3 & 42.2 & 39.67 & 196.1M \\
\midrule
\multirow{5}{*}{RDS+}
& 1K  & 33.3 & 27.7 & 29.8 & 30.27 & 12.1M  \\
& 3K  & 36.2 & 29.8 & 35.4 & 33.80 & 34.8M \\
& 5K  & 38.8 & 33.8 & 36.6 & 36.40 & 56.1M \\
& 10K & 45.0 & 31.9 & 38.8 & 38.57 & 104.9M \\
& 15K & 45.2 & 33.5 & 39.9 & 39.53 & 149.9M \\
\midrule

\multirow{5}{*}{TEMP}
& 1K  & 30.4 & 24.6 & 28.1 & 27.70 & 6.8M  \\
& 3K  & 31.0 & 28.3 & 32.5 & 30.60 & 20.1M \\
& 5K  & 35.4 & 30.4 & 37.9 & 34.57 & 32.0M \\
& 10K & 42.5 & 30.4 & 35.4 & 36.10 & 64.9M \\
& 15K & 42.7 & 33.3 & 38.8 & 38.27 & 96.4M \\
\midrule
\multirow{5}{*}{DeGMatch}
& 1K  & 34.6 & 26.2 & 31.7 & 30.83 & 13.8M  \\
& 3K  & 37.9 & 31.0 & 35.6 & 34.83 & 36.1M  \\
& 5K  & 42.3 & 31.2 & 36.2 & 36.57 & 55.7M  \\
& 10K & 47.7 & 33.8 & 42.6 & \textbf{41.37} & 98.3M  \\
& 15K & 45.4 & 33.8 & 40.0 & 39.73 & 134.1M \\
\bottomrule
\end{tabular}%
}

\caption{
Detailed results across selection budgets. We report performance on AIME24, AIME25, and AIME26, their arithmetic mean, and the total number of training tokens in each selected subset.
}
\label{tab:budget_detailed}
\vspace{-10pt}
\end{table}

\subsection{Detailed Budget-Scaling Results}
\label{app:budget_scaling_details}

Table~\ref{tab:budget_detailed} reports the benchmark-level results underlying the budget-scaling analysis in Section~\ref{sec:budget_scaling}.
For each selection method and budget, we report performance on AIME24, AIME25, and AIME26, their arithmetic mean, and the number of training tokens in the selected subset.
These results correspond to the instance- and token-budget views shown in Figure~\ref{fig:budget_scaling}.

\end{document}